\documentclass[twoside]{article}

\usepackage[accepted]{aistats2021}
\usepackage{amsmath}
\usepackage{amssymb}
\usepackage{bbm}
\usepackage{xcolor}
\usepackage{graphicx}
\usepackage{amsthm}
\usepackage{soul}
\usepackage{booktabs}
\usepackage{fancyvrb}
\usepackage[ruled, vlined, noend]{algorithm2e}
\usepackage{enumitem}
\usepackage{caption}
\usepackage{subcaption}
\usepackage{hyperref}
\usepackage[htt]{hyphenat}
\usepackage{comment}
\usepackage{multirow}
\usepackage{makecell}

\usepackage[round]{natbib}

\DeclareMathOperator*{\argmax}{arg\,max}
\DeclareMathOperator*{\argmin}{arg\,min}

\newcommand{\acq}{\mathcal{A}} 
\newcommand{\DU}{\mathcal{D}^U} 
\newcommand{\DL}{\mathcal{D}^L} 
\newcommand{\Exp}[2][]{\ensuremath{\mathbb{E}_{#1}\left[#2\right]}}                
\newcommand{\defeq}{\overset{\mathrm{def}}{=}}  

\usepackage{xpatch}
\newtheoremstyle{mytheoremstyle}{0.5ex}{0.5ex}{\itshape}{}{\bfseries}{.}{.5em}{}
\theoremstyle{mytheoremstyle}

\newtheorem{proposition}{Proposition}
\newtheorem{definition}{Definition}
\newtheorem*{remark}{Remark}
\makeatletter
\xpatchcmd{\proof}{\topsep6\p@\@plus6\p@\relax}{}{}{}
\makeatother
\newenvironment{proofsketch}{%
  \proof}{\endproof}

\usepackage{mathtools}

\newcommand{\BLOCKCOMMENT}[1]{}

\begin{document}

\runningtitle{Optimal Deep Active Learning Behaviors}

\twocolumn[

\aistatstitle{Towards Understanding the Behaviors of\\Optimal Deep Active Learning Algorithms}

\vspace{-0.18in}

\aistatsauthor{ Yilun Zhou$^*$ \And Adithya Renduchintala$^\dagger$ \And Xian Li$^\dagger$}

\vspace{0.04in}

\aistatsauthor{ Sida Wang$^\dagger$ \And Yashar Mehdad$^\dagger$ \And Asish Ghoshal$^\dagger$}

\vspace{0.04in}

\aistatsaddress{$^*$MIT CSAIL \And $^\dagger$Facebook AI}

\vspace{-0.10in}

\runningauthor{Yilun Zhou, Adithya Renduchintala, Xian Li, Sida Wang, Yashar Mehdad, Asish Ghoshal}

]

\graphicspath{{figures/}}

\begin{abstract}
Active learning (AL) algorithms may achieve better performance with fewer data because the model guides the data selection process. 
While many algorithms have been proposed, there is little study on what the \textit{optimal} AL algorithm looks like, which would help researchers understand where their models fall short and iterate on the design.
In this paper, we present a simulated annealing algorithm to search for this optimal oracle and analyze it for several tasks. We present qualitative and quantitative insights into the behaviors of this oracle, comparing and contrasting them with those of various heuristics. Moreover, we are able to consistently improve the heuristics using one particular insight. We hope that our findings can better inform future active learning research. The code is available at \url{https://github.com/YilunZhou/optimal-active-learning}. 
\end{abstract}

\section{Introduction}
Training deep models typically requires a large dataset, which limits its usefulness in domains where expensive expert annotation is required. Traditionally, active learning (AL), in which the model selects the data points to annotate and learn from, is often presented as a more sample efficient alternative to standard supervised learning \citep{settles2009active}. 
However the gain of AL with deep models is less consistent. For example, the best method seems to depend on the task in an unpredictable manner \citep{lowell2018practical}.  

Is active learning still useful in the era of deep learning? Although many issues are identified, it is not clear whether those problems only plague current AL methods or also apply to methods developed \textit{in the future}. Moreover, in existing literature, there lacks a comparison of proposed methods to the \textit{oracle upper-bound}. Such a comparison is helpful for debugging machine learning models in many settings. 
For example, the classification confusion matrix may reveal the inability of the model to learn a particular class, and a comparison between a sub-optimal RL policy and the human expert play may indicate a lack of exploration. By contrast, without such an oracle reference in AL, it is extremely difficult, if at all possible, to pinpoint the inadequacy of an AL method and improve it. 

In this paper, we propose a simulated annealing algorithm to search for the \textit{optimal} AL strategy for a given base learner. With practical computational resources, this procedure is able to find an oracle that significantly outperforms existing heuristics (+7.53\% over random orders, compared to +1.49\% achieved by the best heuristics on average across three vision and language tasks), for models both trained from scratch and finetuned from pre-training, \textbf{definitively asserting the usefulness of a high-performing AL algorithm in most scenarios (Sec.~\ref{sec:quant-results}).} We also present the following insights into its behaviors. 

While many papers do not explicitly state whether and how training stochasticity (e.g. model initialization or dropout) is controlled across iterations, we show that \textbf{training stochasticity tends to negatively affect the oracle performance (Sec.~\ref{analysis-stochasticity}).}

Previous work \citep{lowell2018practical} has found that for several heuristic methods, the actively acquired dataset does not transfer across different architectures. We observed a lesser extent of this phenomenon, but more importantly, \textbf{the oracle transfers better than heuristics (Sec.~\ref{analysis-transfer}).}

It may seem reasonable that a high-performing AL algorithm should exhibit a non-uniform sampling behavior (e.g. focusing more on harder to learn regions). However, \textbf{the oracle mostly preserves data distributional properties (Sec.~\ref{analysis-distribution}).}

Finally, using the previous insight, \textbf{heuristics can on average be improved by 2.95\% with a simple distribution-matching regularization (Sec.~\ref{analysis-regularization}).}

\section{Related Work}
\label{related-work}

Active learning \citep{settles2009active} has been studied for a long time. At the core of an active learning algorithm is the acquisition function, which generates or selects new data points to label at each iteration. Several different types of heuristics have been proposed, such as those based on uncertainty \citep{kapoor2007active}, disagreement \citep{houlsby2011bayesian}, diversity \citep{xu2007incorporating}, and expected error reduction \citep{roy2001toward}. In addition, several recent studies focused on meta-active learning, i.e. learning a good acquisition function on a source task and transfer it to a target task \citep{konyushkova2017learning, fang2017learning, woodward2017active, bachman2017learning, contardo2017meta, pang2018meta, konyushkova2018discovering, liu2018learning, vu2019learning}. 

With neural network models, active learning has been applied to computer vision (CV) \citep[e.g.][]{gal2017deep, kirsch2019batchbald, sinha2019variational} and natural language processing (NLP) \citep[e.g.][]{shen2017deep, fang2017learning, liu2018learning, kasai2019low} tasks. 
However, across different tasks, it appears that the relative performance of different methods can vary widely: a method can be the best on one task while struggling to even outperform the random baseline on another, with little explanation given. We compile a meta-analysis in App.~\ref{app:meta-analysis}. Moreover, \citet{lowell2018practical} found that the data acquisition order does not transfer well across architectures, a particularly important issue during deployment when it is expected that the acquired dataset will be used for future model development (i.e. datasets outliving models). 

The closest work to ours is done by \citet{koshorek2019limits}, who studied the active learning limit. There are numerous differences. First, we analyze several CV and NLP tasks, while they focused on semantic role labeling (SRL) in NLP. Second, we explicitly account for training stochasticity, shown to be important in Sec.~\ref{analysis-stochasticity}, but they ignored it. Third, our global simulated annealing search is able to find significant gaps between the upper limit and existing heuristics while their local beam search failed to find such a gap (though on a different task). We additionally show how to improve upon heuristics with our insights.

\section{Problem Formulation}
\label{problem-formulation}
\subsection{Active Learning Loop}
Let the input and output space be denoted by $\mathcal{X}$ and $\mathcal{Y}$
respectively, where $\mathcal{Y}$ is a finite set. 
Let $\mathbb P_{XY}$ be the data distribution over $\mathcal{X} \times \mathcal{Y}$.  We consider multi-class classification where a model $m_\theta: \mathcal{X} \rightarrow \mathcal{Y}$, parameterized by $\theta$, maps an input to an output. 

We study pool-based batch-mode active learning
where we assume access to an unlabeled data pool $\DU$ drawn from the
marginal distribution $\mathbb P_{X}$ over $\mathcal{X}$.

Starting with a labeled warm-start set $\DL_0\sim\mathbb P_{XY}$, an AL loop builds a sequence of models $(\theta_0, \ldots, \theta_K)$ and a sequence of labeled datasets $(\DL_1, \ldots, \DL_K)$ in an interleaving manner. 
At the $k$-th iteration, for $k=0, \ldots, K$, a trainer $\eta$ takes in $\DL_k$ and produces the model parameters $\theta_k$ by minimizing some loss function on $\DL_k$. In deep-learning
the model training is typically stochastic (due to, for example, random initialization and drop-out masking) and $\theta_k$ is a random variable. 
We assume that all such stochasticity are captured in $\xi$ such that $\theta_k = \eta(\DL_k, \xi)$ is deterministic.

Using the trained model $m_{\theta_k}$ and the current labeled set $\DL_k$, an acquisition function $\acq$ builds the dataset $\DL_{k+1}$ by selecting a batch of $B$ data points from the unlabeled pool, $\Delta \mathcal D_{k+1}\subseteq \DU$, querying the annotator for labels, and adding them to $\DL_k$; i.e. $\Delta \mathcal D_{k+1} = \acq(\theta_k, \DL_k); \DL_{k+1} = \DL_{k} \cup \Delta \mathcal D_{k+1}$.\footnote{Some acquisition functions such as BALD are stochastic. For simplicity, we discuss the case of deterministic $\acq$ here. Extension of our claims to the stochastic case is straightforward and presented in App.~\ref{app:stochastic-acq}.} The model training and data acquisition loop repeats until we obtain $\DL_K$ by querying labels for $KB$ data points, followed by training $m_{\theta_K}$ for a final time. 

\subsection{Performance Curve}
Given a specific draw of $\xi$, we measure the performance of an acquisition function $\acq$ by its \emph{performance
curve} $\tau_{\acq, \xi}: \{1,\ldots,K\} \rightarrow [0,1]$ defined below:
\begin{align*}
    \tau_{\acq, \xi}(k) &= 
        \Exp[x,y \sim \mathbb P_{XY}]{e(m_{\theta_k}(x), y)}, 
\end{align*}
where $e: \mathcal{Y} \times \mathcal{Y} \rightarrow [0,1]$ is the evaluation metric (e.g. accuracy). $\theta_k = \eta(\DL_k, \xi)$, and $\DL_k = \DL_{k-1}\cup \acq(\theta_{k-1}, \DL_{k-1})$. 

\begin{definition}[Anytime optimality]
An acquisition function $\acq$ is $\xi$-anytime optimal if it uniformly dominates every other
acquisition function $\acq'$ as measured by $\tau_{\cdot, \xi}$; i.e.,
$\tau_{\acq, \xi}(k) \geq \tau_{\acq', \xi}(k)$ $\forall k \in \{1,\ldots,K\}$ and $\forall \acq' \neq \acq$.
\end{definition}

\begin{proposition}
\label{anytime-optimal-proposition}
There exist data distribution $\mathbb P_{XY}$ and model class $m_{\theta}$
for which an anytime optimal acquisition function does not exist.
\end{proposition}
\begin{proofsketch}
Fig.~\ref{fig:counterex} shows a counter-example. In Fig.~\ref{fig:counterex}(a), we have an underlying distribution $\mathbb P_{XY}$ shown as the colored background, and four points drawn from the distribution. If we learn a max-margin linear classifier from the four points, we can uncover the ground-truth decision boundary. If we acquire two data points, the optimal combination is shown in Fig.~\ref{fig:counterex}(b), resulting in a slightly wrong decision boundary. Choosing a different blue point would result in a much worse decision boundary (Fig.~\ref{fig:counterex}(c)). However, the optimal three-point acquisition (Fig.~\ref{fig:counterex}(d)), which leads to the ground-truth decision boundary, does \textit{not} contain the previously optimal blue point in Fig.~\ref{fig:counterex}(b). Thus, there is no acquisition function simultaneously optimal at both two and three data points. 
\begin{figure}[!htb]
    \centering
    \includegraphics[width=\columnwidth]{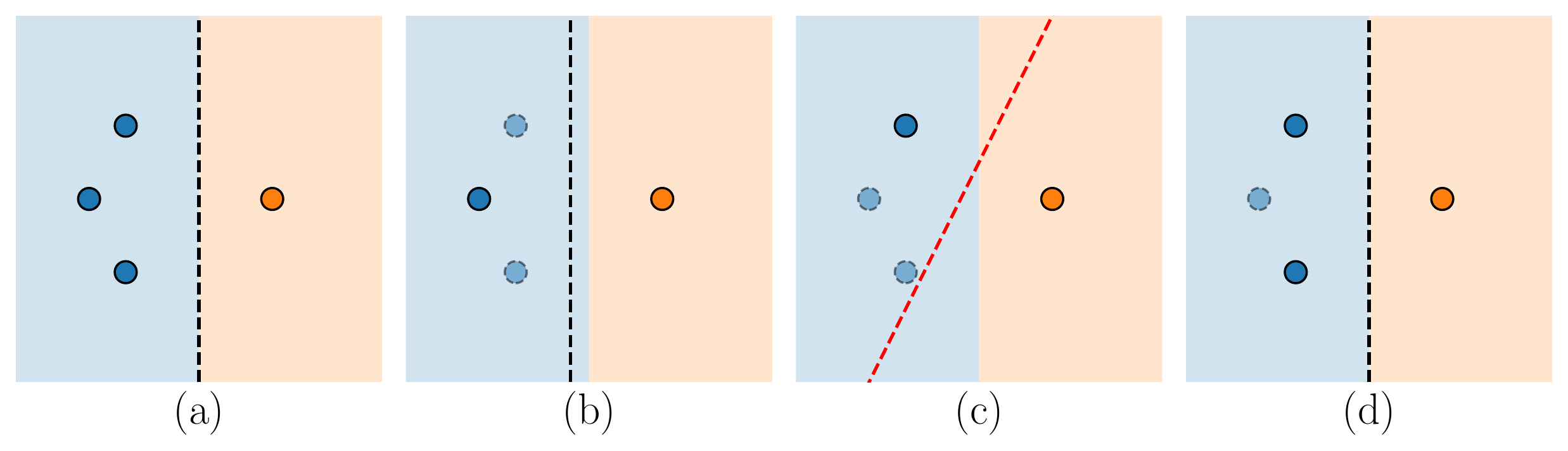}
    \caption{The counter example to prove Prop.~\ref{anytime-optimal-proposition}. }
    \label{fig:counterex}
\end{figure}
\end{proofsketch}

\begin{remark}
In most AL papers, the authors demonstrate the superiority of their proposed method by showing its performance curve visually ``above'' those of the baselines. However, Prop.~\ref{anytime-optimal-proposition} shows that such an anytime optimal acquisition function may not exist. 
\end{remark}

\subsection{$\xi$-Quality}
To make different acquisition functions comparable, we propose the following average performance metric, which we will refer to as \emph{$\xi$-quality}: 
\begin{align*}
    q_{\xi}(\acq) \defeq 
    \frac{1}{K} \sum_{k=1}^{K} \tau_{\acq, \xi}(k). 
\end{align*}
We refer to the acquisition strategy that maximizes this $\xi$-quality as \textit{$\xi$-optimal} strategy, denoted as $\acq_\xi^*$. 
Obviously, the $\xi$-anytime optimal acquisition function, if exists, will also be $\xi$-optimal. 

While we could define the quality simply as the end point performance, $\tau_{\acq, \xi}(K)$, doing so fails to distinguish acquisition functions that lead to a fast initial performance rise from the rest, potentially incurring more annotation cost than necessary. 

There are two interpretations to the above quality definition. First, it is the right Riemann sum approximation of the area under curve (AUC) of $\tau_{\acq, \xi}$, which correlates with intuitive notions of optimality for acquisition functions. Second, the un-averaged version of our definition can also be interpreted as the undiscounted cumulative reward of an acquisition \textit{policy} over $K$ steps where the per-step reward is the resulting model performance following data acquisition. Thus, the optimal policy implicitly trades off between immediate ($k=1$) and future ($k\rightarrow K$) model performances.

\subsection{Optimal Data Labeling Order}
We index data points in $\DU$ by $\DU_{i}$ and use $y\left(\DU_{i}\right)$ to denote the label. Since an acquisition function generates a sequence of labeled datasets
$\DL_1 \subset \ldots \subset \DL_K$, 
an acquisition function is equivalent to a partial permutation order $\sigma$ of $KB$ indices of ${\DU}$ with 
$\DL_k = \DL_0\cup \{\left(\DU_{\sigma_i}, y\left(\DU_{\sigma_i}\right)\right) \}_{i \in [kB]}$. Thus, the problem of searching for $\acq_\xi^*$ reduces to that of searching for the $\xi$-optimal order $\sigma_\xi^*$, which also relieves us from explicitly considering different forms of acquisition functions. 

Moreover, a direct implication of above definitions is that optimal order depends on $\xi$; i.e. $q_\xi(\sigma_\xi^*)\geq q_\xi(\sigma_{\xi'}^*)$, for $\xi\neq \xi'$. As we experimentally demonstrate in Sec.~\ref{analysis-stochasticity}, such a gap does exist. Since stochasticity in model training is completely controllable, an acquisition function that approaches optimal limit may need to explicitly take such randomness into account. 

\begin{table*}[!t]
    \centering
\resizebox{\textwidth}{!}{
    \begin{tabular}{lllrllll}\toprule
        Task & Type & Dataset & $|\DU|, |\DL_0|, |\mathcal D^M|, |\mathcal D^V|, |\mathcal D^T|$ & $B, \,\,K$ & Metric & Architecture & Heuristics\\\midrule
        OC & class. & Fashion-MNIST & 2000, \phantom{0,}50, \phantom{,}150, 4000, 4000 & 25, 12 & Acc & CNN & Max-Ent., (Batch)BALD\\\midrule
        IC & class. & TOPv2 (alarm) & 800, \phantom{0,}40, \phantom{,}100, 4000, 4000 & 20, 8 & F1 & \makecell[cl]{LSTM, CNN, \\AOE, RoBERTa} & Max-Ent., BALD\\\midrule
        NER & tagging & MIT Restaurant & 1000, \phantom{0,}50, \phantom{,}200, 3000, 3000 & 25, 10 & F1 & LSTM & (Norm.-)Min-Conf., Longest\\\bottomrule
    \end{tabular}
}
    \caption{Summary of experiment settings. Architecture details are in App.~\ref{app:arch}. }
    \label{tab:overview}
\end{table*}

\section{Search for the Optimal Order}
\label{search}
There are two technical challenges in finding $\sigma_\xi^*$. First, we do not have access to the data distribution. Second, the permutation space of all orders is too large. 

\subsection{Validation Set Maximum}
To solve the first problem, we assume access to a validation set $\mathcal D^V \sim \mathbb P_{XY}$.  Since we are searching for the oracle model, there is no practical constraints on the size of $\mathcal D^V$. In addition, we assume access to an independently drawn test set $\mathcal D^{T}\sim \mathbb P_{XY}$. We define our optimal order estimator as 
\begin{align}
    \widehat \sigma_{\xi}=\argmax_\sigma q_{\xi}\left(\sigma, \mathcal D^{V}\right), 
\end{align}
and its quality on the test set $q_{\xi}\left(\sigma, \mathcal D^{T}\right)$ serves as the an \textit{unbiased} estimate of its generalization quality. 

\subsection{Simulated Annealing Search}

Exhaustively searching over the space of all labeling orders is prohibitive as there are $|\mathcal D^U|! / \left[(|\mathcal D^U| - BK)!\cdot (B!)^K\right]$ different orders to consider. In addition, we cannot use gradient-based optimization due to the discreteness of order space. Thus, we use simulated annealing (SA) \citep{kirkpatrick1983optimization}, as described in Alg.~\ref{sa}. 

\begin{algorithm}[!htb]
\SetAlgoLined
\SetKwInput{KwInput}{Input}
\KwInput{SA steps $T_S$, greedy steps $T_G$, linear annealing factor $\gamma$}
\hspace{-0.05in}\begin{tabular}{p{0.23cm}l}
$\sigma^{(0)}$ & $=\texttt{random-shuffle}\left([1, 2, ..., |\DU|]\right);$\\
$q^{(0)}$ & $=q_{\xi}(\sigma^{(0)});$\\
$\sigma^*$& $=\sigma^{(0)}; q^*=q^{(0)};$\\
\end{tabular}\\
\For{$t = 1, 2, ..., T_S$}{
    \hspace{-0.07in}\begin{tabular}{p{0.23cm}l}
    $\sigma^{(p)}$ & $=\texttt{propose}\left(\sigma^{(t-1)}\right);$\\
    $q^{(p)}$ & $=q_{\xi}\left(\sigma^{(p)}, \mathcal D^V\right);$\\
    $u$ & $\sim\mathrm{Unif}[0, 1]$;\\
    \end{tabular}\\
    \uIf{$u<\exp\left[\gamma t \left(q^{(p)}-q^{(t-1)}\right)\right]$}{
        $\sigma^{(t)}=\sigma^{(p)}; q^{(t)}=q^{(p)}$;\\
        \If{$q^* < q^{(p)}$}{
            $\sigma^*=\sigma^{(p)}; q^*=q^{(p)}$;\\
        }
    }\Else{
        $\sigma^{(t)}=\sigma^{(t-1)}; q^{(t)}=q^{(t-1)}$;
    }
}
\For{$t = 1, 2, ..., T_G$}{
    \hspace{-0.07in}\begin{tabular}{p{0.23cm}l}
    $\sigma^{(p)}$ & $=\texttt{propose}\left(\sigma^*\right);$\\
    $q^{(p)}$ & $=q_{\xi}\left(\sigma^{(p)}, \mathcal D^V\right);$\\
    \end{tabular}\\
    \uIf{$q^{(p)}>q^*$}{
        $\sigma^*=\sigma^{(p)}; q^*=q^{(p)}$;\\
    }
}

\Return{$\sigma^*, q^*$}
\caption{Simulated Annealing (SA) Search}
\label{sa}
\end{algorithm}

\begin{figure}[!htb]
    \centering
    \vspace{-0.15in}
    \includegraphics[trim={0.05in, 0, 0, 0}, clip, width=0.95\columnwidth]{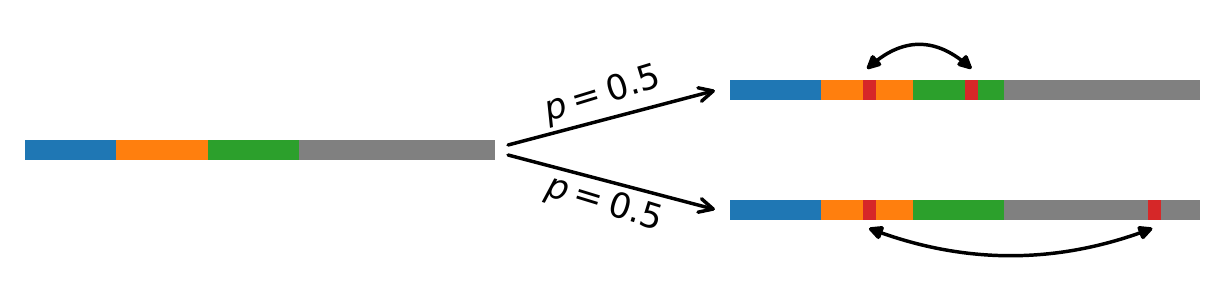}
    \vspace{-0.15in}
    \caption{Illustration of the transition kernel. With equal probabilities, the kernel either swaps two data points from two different batches (sections of different colors) or replaces a data point in the first $K$ batches with one outside (the gray section). }
    \label{fig:swap-kernel}
    \vspace{-0.13in}
\end{figure}

The search starts with a randomly initialized order $\sigma^{(0)}$. At time step $t$, it proposes a new order $\sigma^{(p)}$ from $\sigma^{(t-1)}$ with the following transition kernel (Fig.~\ref{fig:swap-kernel}): with equal probabilities, it either swaps two data points from two different batches or replaces a data point in use by an unused one. It then evaluates the quality of the new order and accepts or rejects the proposal with probability depending on the quality difference. After searching for $T_S$ steps, the algorithm greedily optimizes the best order for an additional $T_G$ steps, and returns best order and quality in the end. 

\section{Experiment Overview}
\label{experiment-overview}

In addition to datasets $\DU, \DL_0, \mathcal D^V$ and $\mathcal D^T$ described in Sec.~\ref{problem-formulation} and \ref{search}, since training neural networks on small datasets is prone to overfitting, we introduced an additional model selection set $\mathcal D^M$ to select the best model across 100 epochs and also trigger early stopping if the performance on this set is not improved for 20 consecutive epochs. In typical AL settings where data annotation is costly, $D^M$ is typically comparable in size to the warm-start set $\DL_0$. 

For each task, Tab.~\ref{tab:overview} describes the experiment-specific parameters. The training used the batch size equal to the acquisition batch size $B$ and the Adam optimizer \citep{kingma2014adam}. Following standard practice, the learning rate is $10^{-3}$ except for RoBERTa finetuning, which is $10^{-4}$. We searched for $T_S=25,000$ and $T_G=5,000$ steps with $\gamma=0.1$. We found little improvement in the greedy stage. 

The full dataset is shuffled and split into non-overlapping sets $\DU, \DL_0, \mathcal D^M, \mathcal D^V, \mathcal D^T$. Since the shuffling is \textit{not} label-stratified, the empirical label distribution in any set (and $\DL_0$ in particular) may not be close to the actual distribution. We made this choice as it better reflects the actual deployment. For each task we considered commonly used architectures. 

\subsection{Object Classification (OC)}
For object classification (OC), we used the \texttt{Fashion-MNIST} dataset \citep{fashionmnist} with ten classes. Even though the dataset is label-balanced, $|\DL_0|=50$ means that the initial warm-start set can be extremely imbalanced, posing additional challenge to uncertainty estimation used by many heuristics. We used a CNN architecture with two convolution layers followed by two fully connected layers. We compared against max-entropy, BALD \citep{gal2017deep}, and BatchBALD \citep{kirsch2019batchbald} heuristics. 

\subsection{Intent Classification (IC)}
For intent classification (IC), we used the Task-Oriented Parsing v2 (\texttt{TOPv2}) dataset \citep{xilun2020low}, which consists of eight domains of human interaction with digital assistants, such as \emph{weather} and \emph{navigation}. We used the ``alarm'' domain for our experiments. In intent classification, given an user instruction such as ``set an alarm at 8 am tomorrow'', the task is to predict the main intent of the sentence (\texttt{create-alarm}). There are seven intent classes: \texttt{create-alarm}, \texttt{get-alarm}, \texttt{delete-alarm}, \texttt{silence-alarm}, \texttt{snooze-alarm}, \texttt{update-alarm}, and \texttt{other}. The dataset is very imbalanced, with \texttt{create-alarm} accounting for over half of all examples. Hence, we used multi-class weighted F1 score as the evaluation metric. 

Our main architecture of study is the bi-directional LSTM (BiLSTM) architecture with word embeddings initialized to GloVe \citep{pennington2014glove}. To study the model transfer quality (Sec.~\ref{analysis-transfer}), we also considered a CNN architecture, which uses 1D convolution layers to process a sentence represented as its word embedding sequence, an Average-of-Embedding (AOE) architecture, which uses a fully connected network to process a sentence represented by the average embedding of the words, and finetuning from the per-trained RoBERTa model \citep{liu2019roberta}. Detailed architectures are described in App.~\ref{app:arch}. We considered max-entropy and BALD heuristics. 

\subsection{Named Entity Recognition (NER)}

Named entity recognition (NER) is a structured prediction NLP task to predict a tag type for each word in the input sentence. We used the \texttt{MIT Restaurant} dataset \citep{liu2013asgard}, which consists of restaurant-related sentences tagged in Beginning-Inner-Outer (BIO) format. Tags include \texttt{amenity}, \texttt{price}, \texttt{location}, etc. For example, the tags for ``\textit{what restaurant near here serves pancakes at 6 am}'' are \texttt{[O, O, B-location, I-location, O, B-dish, O, B-hours, I-hours]}. The outer tag accounts for more than half of all tags, and tag-level weighted F1 score is used as the evaluation metric. 

We used a BiLSTM encoder and an LSTM decoder following \citet{shen2017deep} but without the CNN character encoder as an ablation study does not find it beneficial. We used teacher-forcing at training time and greedy decoding at prediction time. During active acquisition, we assumed that the annotation cost is the same for each sentence regardless of its length. We considered several heuristics. The min-confidence heuristic selects the sentence with the lowest log joint probability of the greedily decoded tag sequence (i.e. sum of log probability of each tag), which is divided by sentence length in its normalized version. The longest heuristic prioritizes longer sentences.

\section{Results and Analyses}
\label{analysis}

We compared the performance of the oracle order 
against existing heuristics. We called the oracle order and performance ``optimal'' even
though they are estimates of the truly optimal ones due to our use
of the validation set for estimating the quality and the simulated annealing search.

\subsection{Optimal Quality Values and Curves}
\label{sec:quant-results}
Tab.~\ref{tab:aucs} presents the qualities of the optimal, heuristic and random orders. On all three tasks, there are large gaps between the heuristic and optimal orders. Specifically, the optimal orders are better than the best heuristics by 11.7\%, 3.1\%, and 3.5\% for OC, IC, and NER respectively in terms of $\xi$-quality. For OC and IC, we repeated the training across five random seeds. The difference between the optimal and heuristic performances is significant at $\alpha=0.05$ for both tasks under a paired $t$-test, with $p\leq 0.001$ and $p=0.003$ respectively. In addition, the optimal orders are better than the random orders by 7.53\% on average across three tasks, while the best heuristic orders only achieves an improvement of 1.49\%.

\begin{figure}[!t]
    \centering
    \includegraphics[width=0.95\columnwidth]{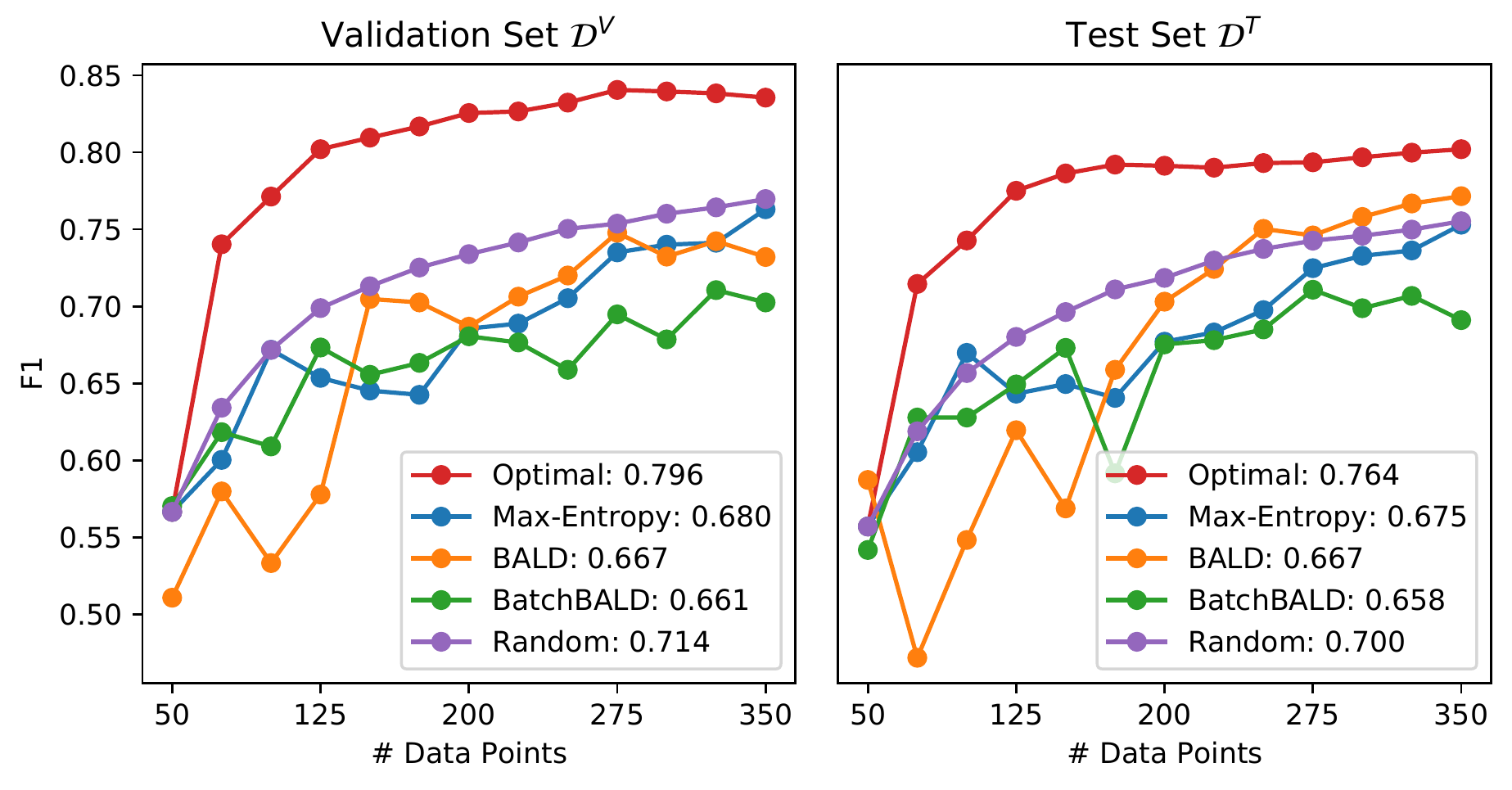}
    \caption{Performance curves of optimal, heuristic, and random orders for object classification. }
    \label{fig:oc-curves}
\end{figure}

Visually, Fig.~\ref{fig:oc-curves} depicts the performance curve of the optimal order against heuristic and random baselines on both the validation
($\mathcal D^V$) and test ($\mathcal D^T$) set for OC. The optimal order significantly outperforms heuristic and random baselines, both numerically and visually, and we observed that the optimal order found using the validation set generalizes to the test set. Quality summary for each individual heuristic and performance curves for IC and NER tasks are presented in App.~\ref{app:qualities-full} and \ref{app:curves}. 
Even though there might not exist an anytime optimal acquisition function (Prop.~\ref{anytime-optimal-proposition}), 
we did observe that the oracle order is uniformly better than
heuristics on all three tasks.

\begin{table}[!t]
    \centering
    \begin{tabular}{llll}\toprule
         & OC & IC & NER \\\midrule
        Optimal & \textbf{0.761} & \textbf{0.887} & \textbf{0.839} \\
        Best Heuristic & 0.682$^*$ & 0.858 & 0.811 \\
        Random & 0.698$^*$ & 0.816 & 0.800\\\bottomrule
    \end{tabular}
    \caption{Qualities of optimal, heuristic, and random orders on the three tasks. Individual heuristic performances are shown in App.~\ref{app:qualities-full}. $^*$No heuristics outperform the random baseline on object classification. }
    \label{tab:aucs}
\end{table}

\subsection{Effect of Training Stochasticity}
\label{analysis-stochasticity}
As previously stated, training stochasticity could negatively affect the optimal quality. To experimentally verify this, for $\xi$ and $\xi'$ we compared $q_{\xi}(\widehat\sigma_{\xi}, \mathcal D^T)$ and $q_{\xi}(\widehat\sigma_{\xi'}, \mathcal D^T)$ to study whether $\xi'$-optimal order is less optimal for $\xi$-training. Since we did not have dropout layers in our network, the only source of stochasticity comes from the random initialization. 

\begin{figure}[!htb]
    \centering
    \includegraphics[width=\columnwidth]{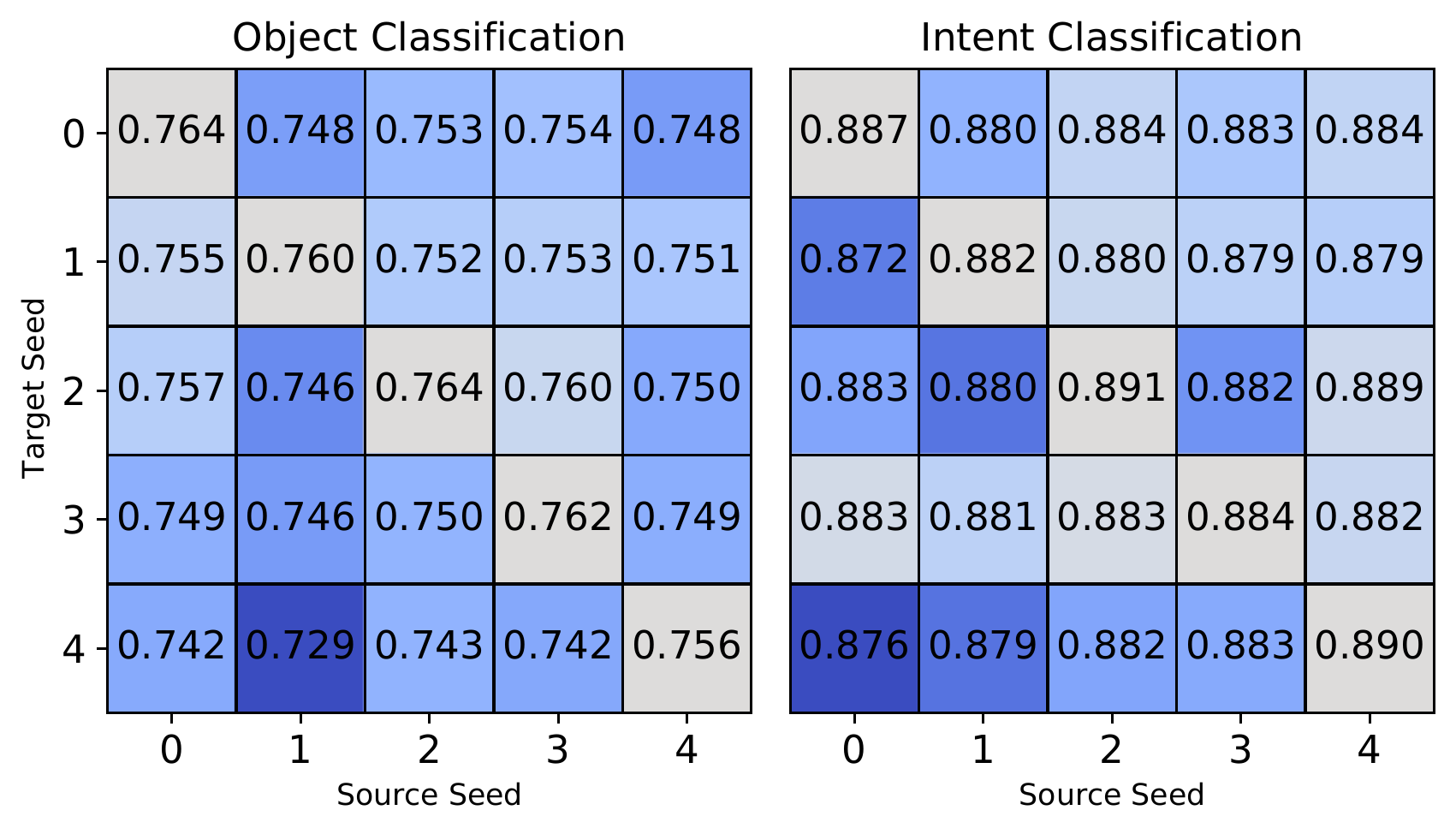}
    \caption{Training stochasticity negatively affects the optimal quality. The number in each cell represents $q_{\xi}(\widehat\sigma_{\xi'})$, with $\xi$ on the row and $\xi'$ on the column. The color represents $q_{\xi}(\widehat\sigma_{\xi'}) - q_{\xi}(\widehat\sigma_{\xi})$, with darker colors indicating larger gaps. 
    }
    \label{fig:seed-mismatch}
\end{figure}

For five different seeds in OC and IC, $\xi, \xi'\in\{0, ..., 4\}$, Fig.~\ref{fig:seed-mismatch} shows the pairwise quality $q_{\xi}(\widehat\sigma_{\xi'}, \mathcal D^T)$, where $\xi'$ is the source seed on the column and $\xi$ is the target seed on the row. The color on each cell represents the seed-mismatch quality gap $q_{\xi}(\widehat\sigma_{\xi}, \mathcal D^T)-q_{\xi}(\widehat\sigma_{\xi'}, \mathcal D^T)$, with a darker color indicating a larger gap. 

The results suggest that in order to fully realize the potential of active learning, the acquisition function needs to specifically consider the particular model. However, high-quality uncertainty estimation are most commonly defined for a \textit{model class} (e.g. the class of all random initialization in Deep Ensemble \citep{lakshminarayanan2017simple}, the class of dropout masks in MC-Dropout \citep{gal2016dropout}, and the class of Gaussian noises in Bayes-by-Backprop \citep{blundell2015weight}). A definition and algorithms for single-model uncertainty have the potential of further improving uncertainty-based heuristics. In addition, the result also implies that a purely diversity-based method relying on pre-defined embedding or distance metric could not be optimal. 

\subsection{Model Transfer Quality}
\label{analysis-transfer}
\citet{lowell2018practical} found that the heuristic acquisition order on one model sometimes does not transfer well to another model. However, it is not clear whether this is a general bane of active learning or just about specific heuristics. On the one hand, if the optimal order finds generally valuable examples early on, we would expect a different model architecture to also benefit from this order. On the other hand, if the optimal order exploits particularities of the model architecture, we would expect it to have worse transfer quality. 

\begin{figure}[!t]
    \centering
    \includegraphics[width=\columnwidth]{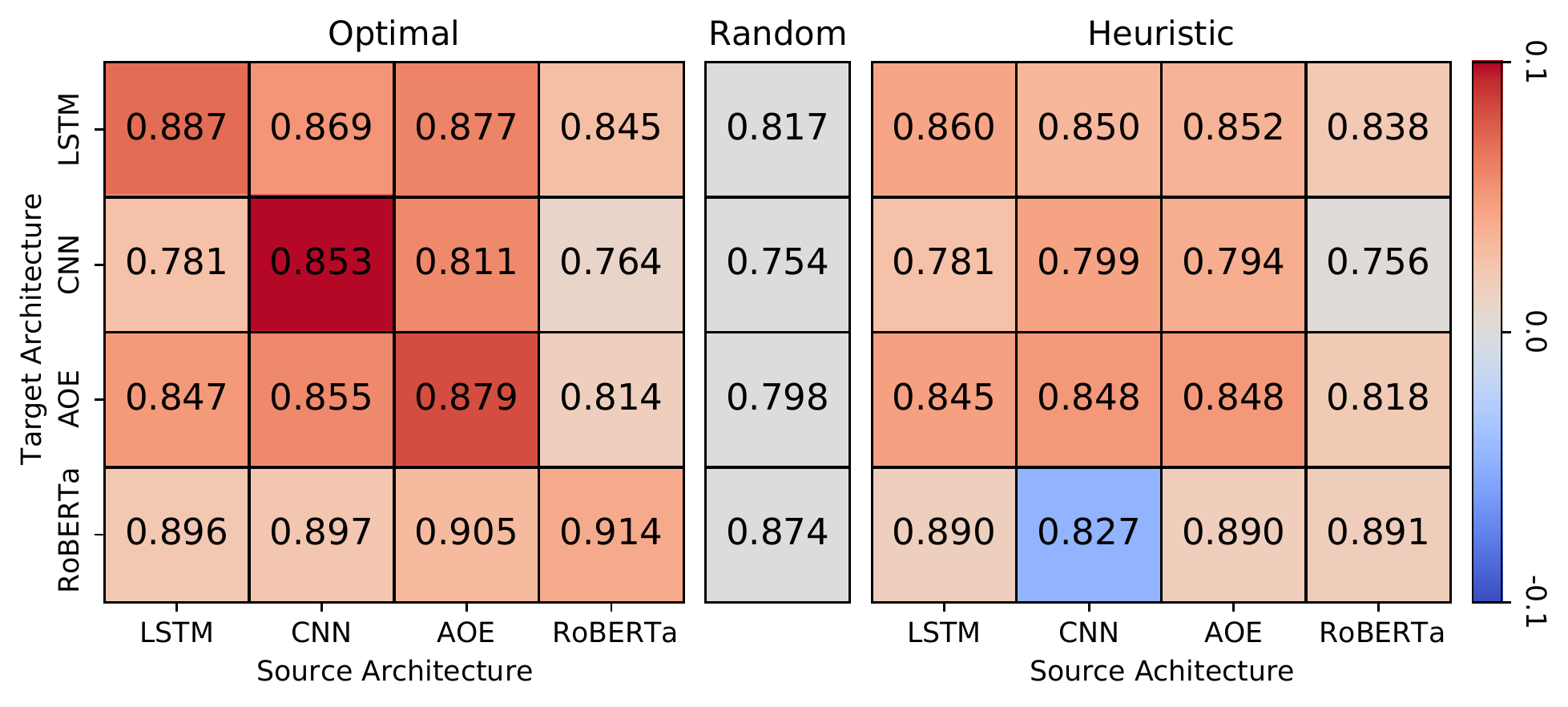}
    \caption{Quality of optimal and max-entropy heuristic order across different architectures. Cell color represents difference to the random baseline. }
    \label{fig:model-transfer}
\end{figure}

We replicated this study on the IC task, using four very different deep architectures: BiLSTM, CNN, AOE and RoBERTa described in Sec.~\ref{experiment-overview}. For a pair of source-target architectures $(A_s, A_t)$, we found both the optimal and best heuristic order on $A_s$ and evaluate it on $A_t$. Then we compared its quality to that of a random order on $A_t$. The best heuristic is BALD for LSTM, CNN and RoBERTa, and max-entropy for AOE. Fig.~\ref{fig:model-transfer} shows the quality of each order. The optimal order always transfers better than the random order and, with the single exception of RoBERTa $\rightarrow$ AOE, better than or equal to the heuristic order as well. This suggests that the optimal order tends to label model-agnostically valuable data points, and it is likely that better acquisition functions for one architecture can also help assemble high-quality datasets for others. 

\begin{figure*}[!htb]
    \centering
    \includegraphics[width=0.85\textwidth]{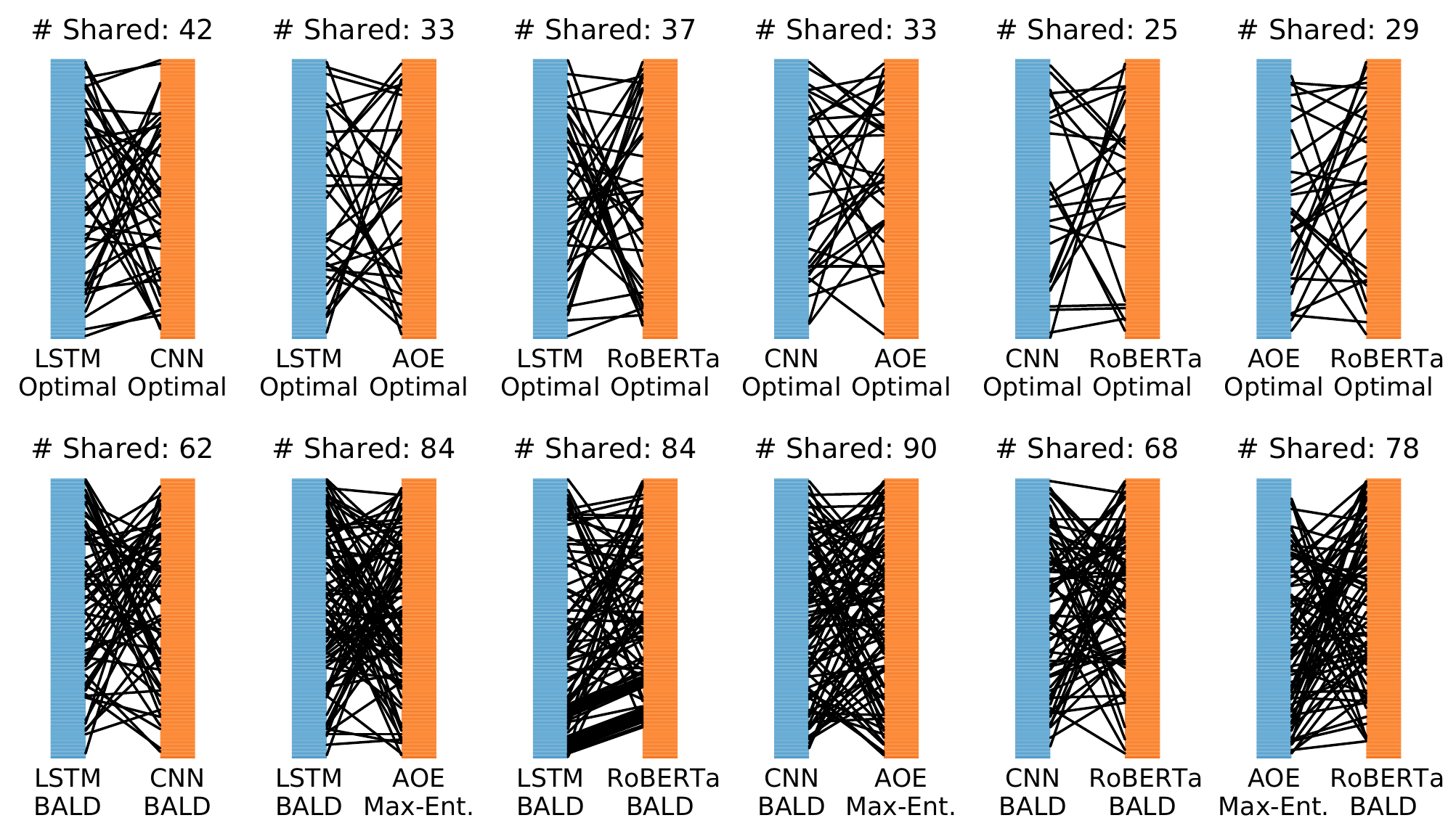}
    \caption{A visual comparison of optimal (top) and heuristic (bottom) orders, for every pair of architectures, showing the number of shared data points acquired under both architectures and their relative ranking. }
    \label{fig:relative-order}
\end{figure*}

Perhaps surprisingly, we found that the optimal orders for different architectures actually share fewer data points than the heuristic orders, despite achieving higher transfer performance. Fig.~\ref{fig:relative-order} shows how the optimal (top) and heuristic (bottom) orders for different architectures relate to each other on the intent classification task. Specifically, each vertical bar is composed of 160 horizontal stripes, which represent the 160 data points acquired under the specified order, out of a pool of 800 data points (not shown). The first stripe from the top represents the first data point being acquired, followed by data points represented by the second, third, etc. stripes from the top. A line between the two orders connects the same data point in both orders and reflects the position of that data point in each of the two orders. The number of data points shared by both orders (i.e. number of connecting lines) is computed and shown. 

\begin{figure*}[!htb]
    \centering
    \includegraphics[width=\textwidth]{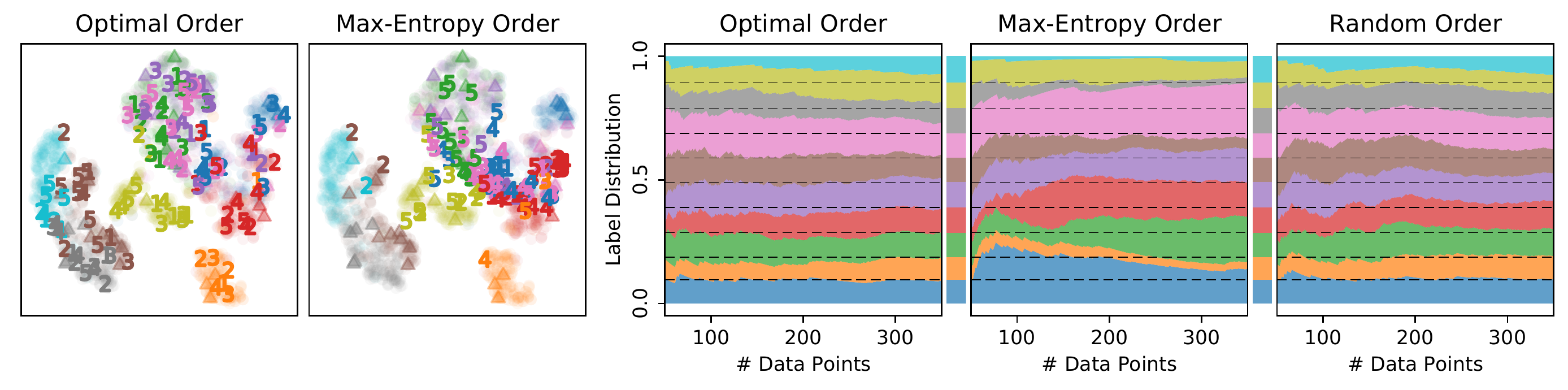}
    \caption{Distribution visualization for object classification. Left: first five batches numerically labeled in $t$-SNE embedding space, along with warm-start (triangle) and test (circle) samples. Right: label distribution w.r.t. labeled set size, with test-set distribution shown between plots and as dashed lines. }
    \label{fig:fmnist-distribution}
\end{figure*}

The results suggest that there are an abundant amount of high quality data points that could lead to high AL performance, even though the optimal order for different architectures selects largely different ones. However, they are mostly missed by the heuristics. The heavy sharing among heuristics can be explained by the fact that they mostly over-sample sentences that are short in length or belong to the minority class, and thus deplete the pool of such examples.

\subsection{Distributional Characteristics}

\label{analysis-distribution}

In this set of experiments, we visualized the distributional characteristics of the acquired data points under the optimal and heuristic orders. We analyzed both the input and the output space. 

\subsubsection{Object Classification}

The left two panels of Fig.~\ref{fig:fmnist-distribution} visualizes the first five batches (125 data points in total) acquired by the optimal and max-entropy orders. For $28\times 28$ grayscale Fashion-MNIST images, we reduced their dimension to 100 using principal component analysis (PCA), then to 2 using $t$-distributed stochastic neighbor embedding ($t$-SNE) \citep{maaten2008visualizing}. Each acquired image, positioned at its 2-dimensional embedding and color-coded by label, is represented by a number referring to the batch index. Circle and triangle marks represent points in the test and warm-start sets. 

The right three panels of Fig.~\ref{fig:fmnist-distribution} visualizes label distribution of the $\DL_k$ during acquisition. The horizontal axis represents the total number of data points (starting from 50 images in $\DL_0$, each order acquires 300 from the pool), while the relative length of each color represents the frequency of each label. The bars between plots show the ``reference distribution'' from the test set, which is balanced in this task. 

For both input and output, the optimal order samples quite uniformly in the sample space, and is not easily distinguishable from random sampling. However, this order is distinctively non-random because it achieves a much higher quality than the random order (Tab.~\ref{tab:aucs}). By contrast, the max-entropy sampling is very imbalanced in both input and output space. BALD and BatchBALD exhibit similar behaviors (App.~\ref{app:oc-distr}). 

\begin{figure*}[!htb]
    \centering
    \includegraphics[width=0.9\textwidth]{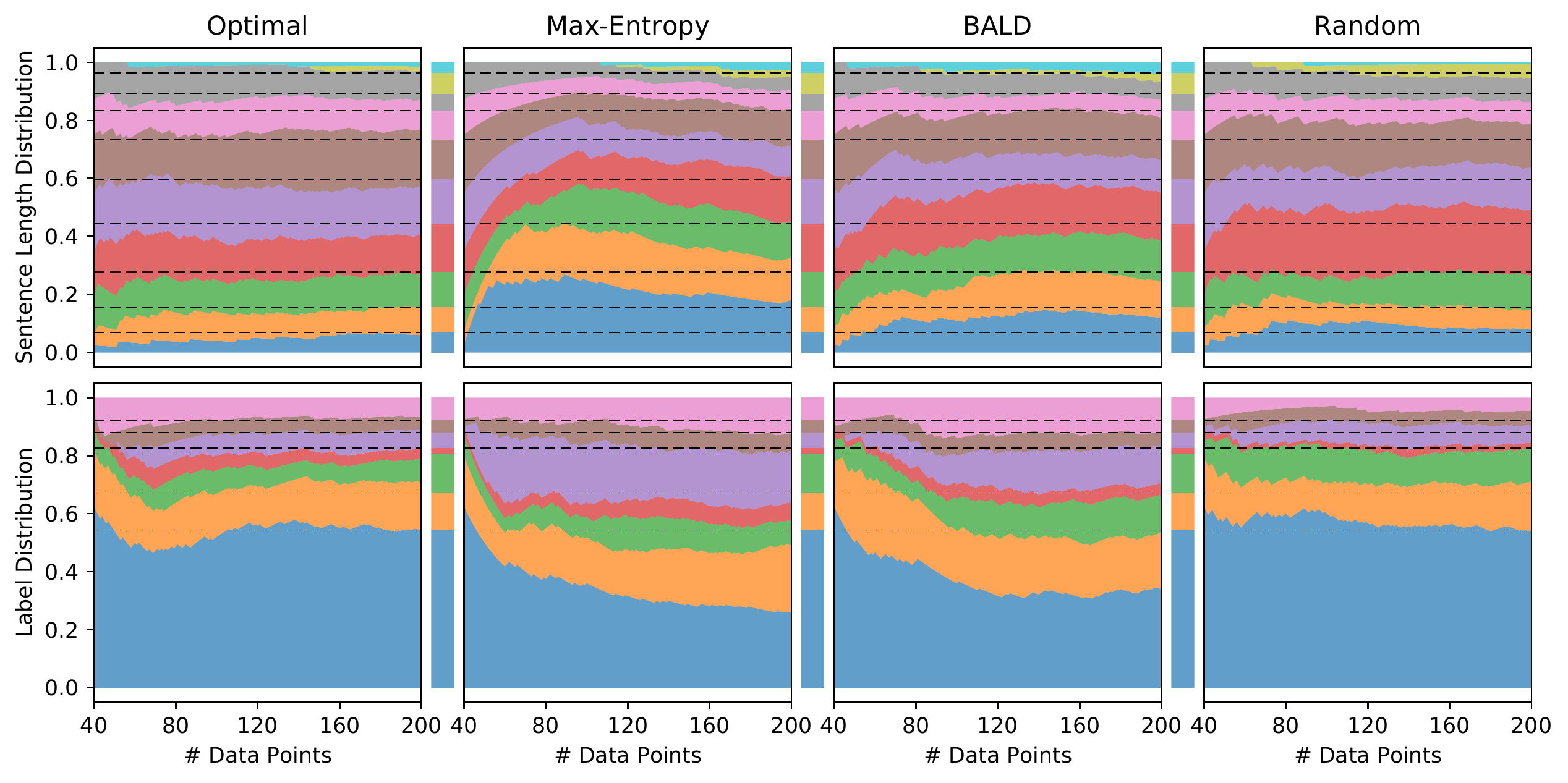}
    \caption{Sentence length and label distribution w.r.t. labeled set size for intent classification. For sentence lengths, the colors represent, from bottom to top: 1--3, 4, 5, 6, 7, 8, 9, 10, 11--12, and 13+. For labels, the colors represent, from bottom to top: \texttt{create-alarm}, \texttt{get-alarm}, \texttt{delete-alarm}, \texttt{other}, \texttt{silence-alarm}, \texttt{snooze-alarm}, and \texttt{update-alarm}. 
    }
    \label{fig:ic-distribution}
\end{figure*}

\begin{figure*}[!htb]
    \centering
    \includegraphics[width=\textwidth]{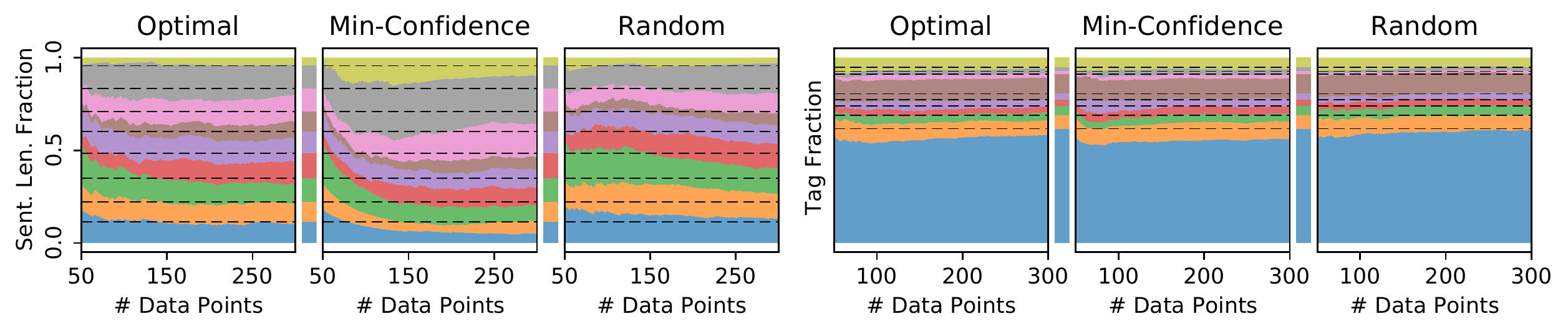}
    \caption{Sentence length and tag distribution w.r.t. labeled set size for NER. For sentence lengths, the colors represent, from bottom to top: 1--5, 6, 7, 8, 9, 10, 11--12, 13-16, and 17+. For labels, the colors represent, from bottom to top: \texttt{Outer}, \texttt{B/I-amenity}, \texttt{B/I-cuisine}, \texttt{B/I-dish}, \texttt{B/I-hours}, \texttt{B/I-location}, \texttt{B/I-price}, \texttt{B/I-rating}, and \texttt{B/I-restaurant-name}. }
    \label{fig:ner-distribution}
\end{figure*}

\begin{figure*}[t]
    \centering
    \vspace{-0.10in}
    \includegraphics[width=0.80\textwidth]{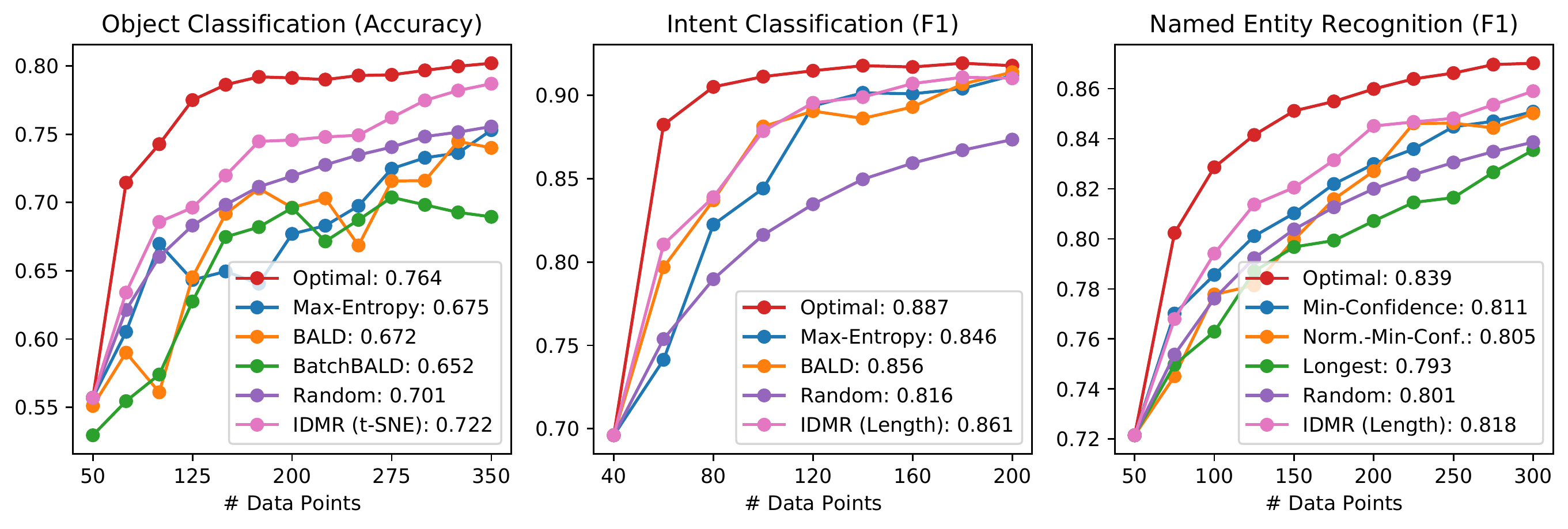}
    \vspace{-0.05in}
    \caption{IDMR-augmented heuristics performance, improving upon the vanilla versions by 2.95\% on average. }
    \vspace{-0.10in}
    \label{idmr-curves}
\end{figure*}

\subsubsection{Intent Classification}

Fig.~\ref{fig:ic-distribution} presents the analogous analysis for IC, where we studied the sentence-length aspect of the input distribution. Since there are few sentences with less than four or more than ten words, we grouped those sentences into 0-3, 11-12, and 13+ categories for ease of visualization. Both input and output distribution plots echo those in Fig.~\ref{fig:fmnist-distribution}. 

In particular, even though short sentences (blue and orange) are infrequent in the actual data distribution, both heuristics over-samples them, likely because shorter sentences provide less ``evidence'' for a prediction. In addition, they also significantly under-samples the \texttt{create-alarm} label (blue), likely because it is already well-represented in the warm-start set, and over-samples the \texttt{silence-alarm} label (purple), for the converse reason. The optimal order, again, does not exhibit any of these under- or over-sampling behaviors while achieving a much higher quality.

\subsubsection{Named Entity Recognition}

Fig. \ref{fig:ner-distribution} presents the analysis for the NER task. We observed the save distribution-matching property, on both the input and the output side. By contrast, even if the acquisition function does not have the ``freedom'' to choose sentences with arbitrary tags, the min-confidence heuristic still over-samples the relatively infrequent \texttt{B/I-hours} tags (purple) in the beginning. 

Notably, in NER the basic ``supervision unit'' is an individual word-tag pair. Thus, the model can effectively learn from more data by selecting longer sentences. This is indeed exploited by the min-confidence heuristic, where very long sentences are vastly over-sampled. However, the optimal order does not exhibit such length-seeking behavior, and instead matches the test set distribution quite closely.

\subsection{Distribution-Matching Regularization}
\label{analysis-regularization}

Sec.~\ref{analysis-distribution} consistently suggests that unlike heuristics, the optimal order matches the data distribution meticulously. Can we improve heuristics with this insight?

\begin{algorithm}[!htb]
\SetAlgoLined
    \SetKwInput{KwInput}{Input}
    \KwInput{$\acq\left(m_\theta, \mathcal D^L, \DU\right)$ that returns the next data point in $\DU$ to label}
    \hspace{-0.01in}\begin{tabular}{p{0.24cm}l}
    $ d_{\mathrm{ref}}$ & $=\texttt{bin-distribution}\left(\DL_{0,X}\cup \DU_X\cup \mathcal D^M\right)$\;\\
    $ d_{\mathrm{cur}}$ & $= \texttt{bin-distribution}\left(\mathcal D_X^L\right)$\;\\
    $ b^*$ & $= \argmin_{b} \left(d_{\mathrm{cur}}-d_{\mathrm{ref}}\right)_b$\;\\
    $ \DU_{b^*}$ & $= \{x\in \DU_X: \texttt{bin}(x)=b^*\}$\;\\
    \end{tabular}\\
    \Return $\acq\left(m_\theta, \mathcal D^L, \DU_{b^*}\right)$
\caption{Input Dist.-Matching Reg. (IDMR)}
\label{idmr}
\end{algorithm}

Alg.~\ref{idmr} presents Input Distribution-Matching Regularization (IDMR), a meta-algorithm that augments any acquisition function to its input distribution-matching regularized version. $\texttt{bin}(x)$ is a function that assigns each input data point $x$ to one of finitely many bins by its characteristics. $\texttt{bin-distribution}(\mathcal D_X)$ computes the empirical count distribution of the bin values for input part of $\mathcal D_X$. The IDMR algorithm compares the reference bin distribution on all available data (i.e. $\DL_{0,X}\cup \DU_X\cup \mathcal D^M_X$) to the current bin distribution of $\mathcal D^L_X$, and uses the base acquisition function to select a data point in the highest deficit bin.

For OC, a $K$-means algorithm identified five clusters in the $t$-SNE space, which are used as the five bins. For IC and NER, sentences are binned by length. We used IDMR to augment the best performing heuristics. As shown in Fig.~\ref{idmr-curves}, on all tasks, the IDMR-augmented heuristic is better than all but the optimal order, with an average improvement of 2.95\% over the its non-augmented counterpart. Across five random seeds for OC and IC, paired $t$-tests yield $p=028$ and $0.017$ for improvement, both significant at $\alpha=0.05$.

We also tried using a similar technique to match the output label distribution. However, it is complicated by the fact that the labels are not observed prior to acquisition and we observed mixed results. The algorithm and results are described in App.~\ref{app:odmr}.

\section{Discussion and Conclusion}
\label{conclusion}

In this paper, we searched for and analyzed the optimal data acquisition order in active learning. Our findings are consistent across tasks and models. First, we confirmed the dependence of optimal orders on training stochasticity such as model initialization or dropout masking, which should be explicitly considered as AL methods approach the optimal limit. 

Notably, we did not find any evidence that the optimal order needs to over-/under-sample in hard-/easy-to-learn regions. 
Given this observation, it is reasonable that the optimal order transfers across architectures better than heuristics and that heuristics can benefit from a distribution-matching regularization. 

Indeed, most AL heuristics seek to optimize proxy objectives such as maximal informativeness or input space coverage. While intuitive, they have not been rigorously justified to correlate with any AL performance (e.g. $\xi$-quality). For example, even if a data point is maximally informative, it is possible that the information could not be fully absorbed during training and lead to performance improvement. 

Moreover, in supervised learning, the fundamental assumption that the training and test data come from the same underlying distribution is the key to most guarantees of empirical risk minimization (ERM). Thus, we conjecture that the distribution-matching property arises from the nice behavior of ERM under this assumption. This also suggests that when the proposed algorithm is expected to violate this assumption, more careful analysis should be done on how such distribution shift may (adversely) affect the performance of models optimized under ERM. 
Developing theoretical guarantees of ERM under controlled distribution mismatch and/or formulations beyond ERM may also benefit active learning. 

Finally, we focused on classification problems with a few classes. It would be interesting for future work to extend the analysis to other settings, such as large number of classes (e.g. CIFAR-100) or regression, and study whether the same behaviors hold or not.

\section*{Acknowledgement}
Yilun Zhou conducted this work during an internship at Facebook AI. Correspondence should be sent to Yilun Zhou at \url{yilun@mit.edu}. We thank the anonymous reviewers for their feedback. 

\bibliographystyle{apalike}
\bibliography{references}

\onecolumn
\clearpage
\appendix

\section{Meta-Analysis of Existing Results}
\label{app:meta-analysis}
\begin{figure}[!htb]
    \centering
    \includegraphics[height=0.35\columnwidth]{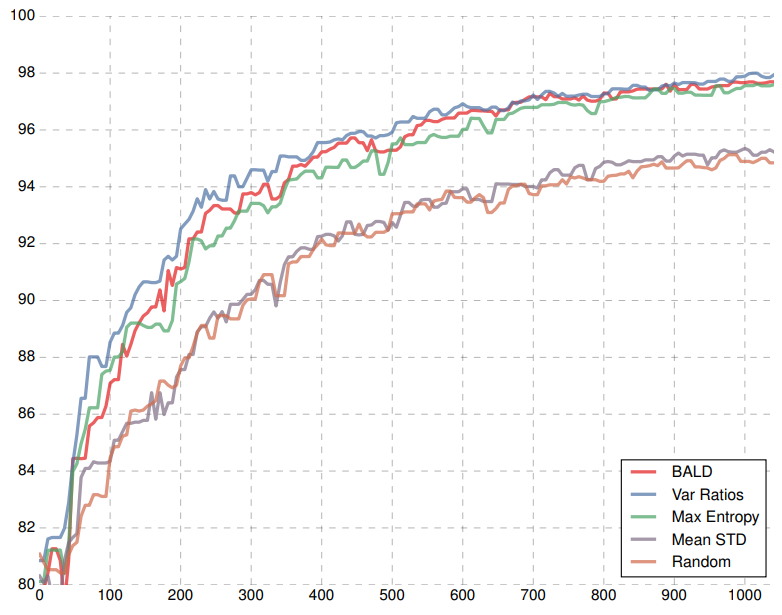}\hspace{0.3in}
    \includegraphics[height=0.35\columnwidth]{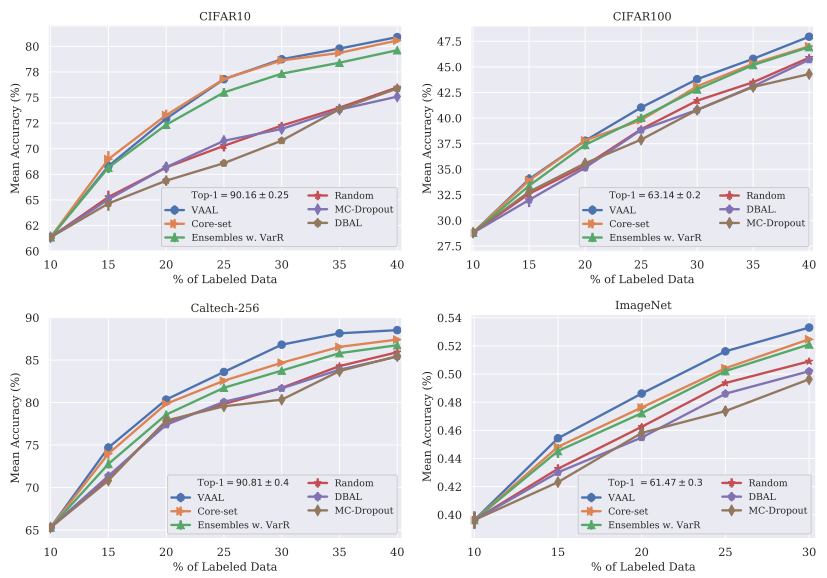}
    \caption{Left: Fig.~1 of \citep{gal2017deep}. Right: Fig.~2 of \citep{sinha2019variational}. }
    \label{fig:meta-figs-1}
\end{figure}

In this section, we present a meta-analysis of existing results to highlight the inconsistencies among them. Fig.~1 of \citep{gal2017deep} (replicated as Fig.~\ref{fig:meta-figs-1} left) shows that BALD (Bayesian active learning by disagreement) using MC-dropout achieves much higher performance than the random baseline on one image dataset. By contrast, Fig.~2 of \citep{sinha2019variational} (replicated as Fig~.\ref{fig:meta-figs-1} right) shows that MC-Dropout methods struggle to outperform even the random baseline on four datasets. 

\begin{figure}[!htb]
    \centering
    \includegraphics[width=0.8\columnwidth]{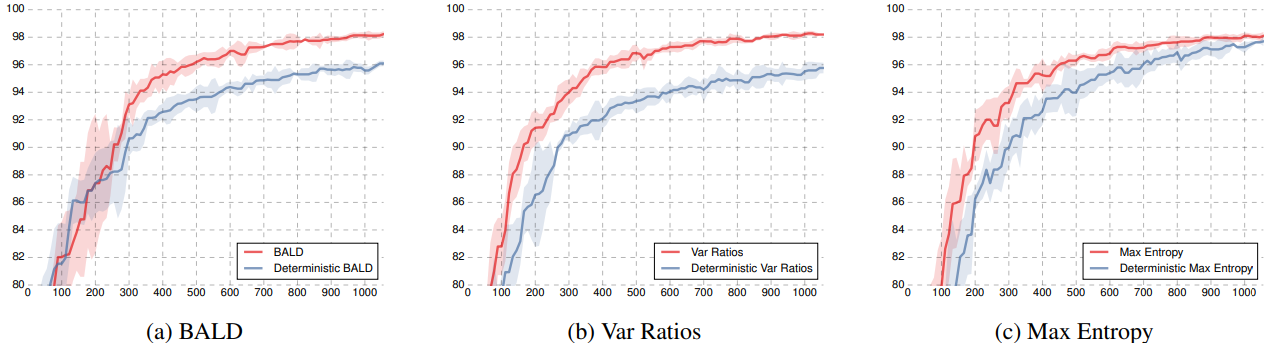}
    \includegraphics[width=0.6\columnwidth]{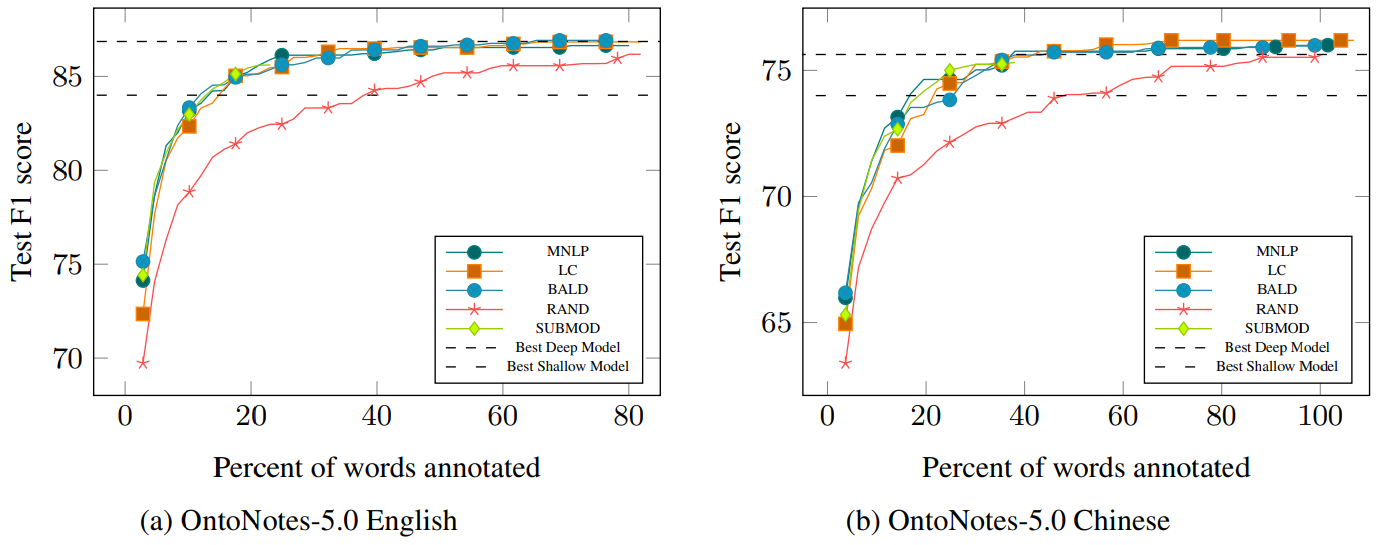}
    \caption{Top: Fig.~2 of \citep{gal2017deep}. Bottom: Fig.~4 of \citep{shen2017deep}. }
    \label{fig:meta-figs-2}
\end{figure}

Fig.~2 of \citep{gal2017deep} (replicated as Fig.~\ref{fig:meta-figs-2} top) shows that active learning strategies with MC-Dropout-based uncertainty estimation consistently outperforms their deterministic counterparts on a CV task. However, Fig.~4 of \citep{shen2017deep} (replicated as Fig.~\ref{fig:meta-figs-2} bottom) finds no discernible difference between MC-Dropout-based BALD and other deterministic heuristics on an NLP task. 

\begin{figure}[!htb]
    \centering
    \includegraphics[width=0.9\columnwidth]{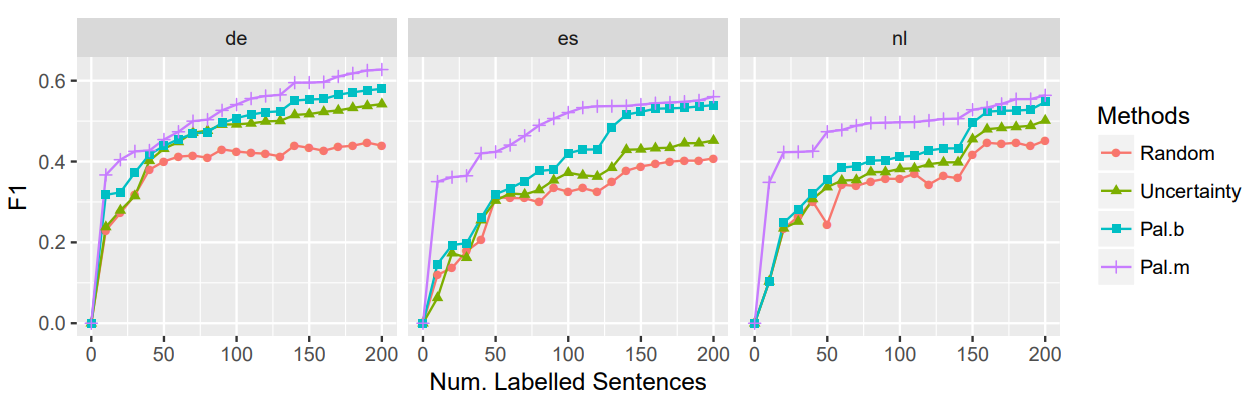}
    \includegraphics[width=0.9\columnwidth]{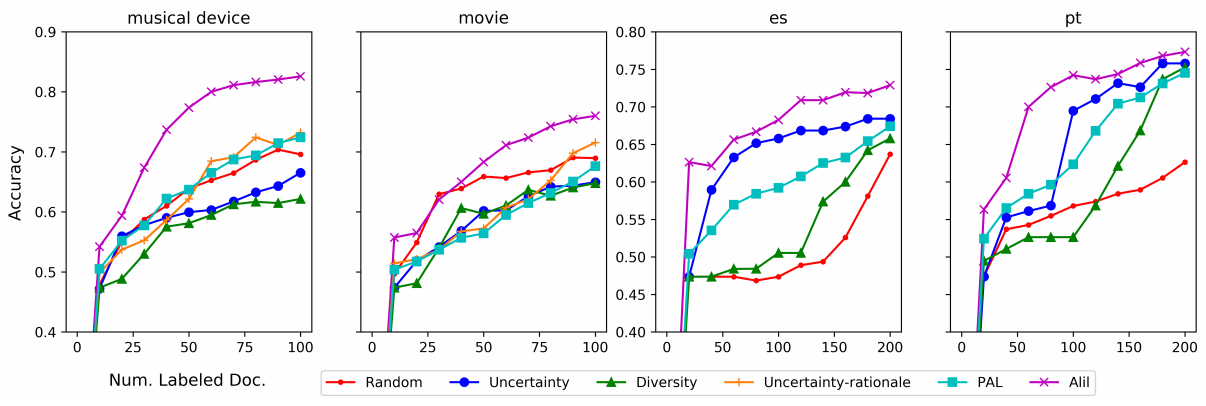}
    \includegraphics[width=0.4\columnwidth]{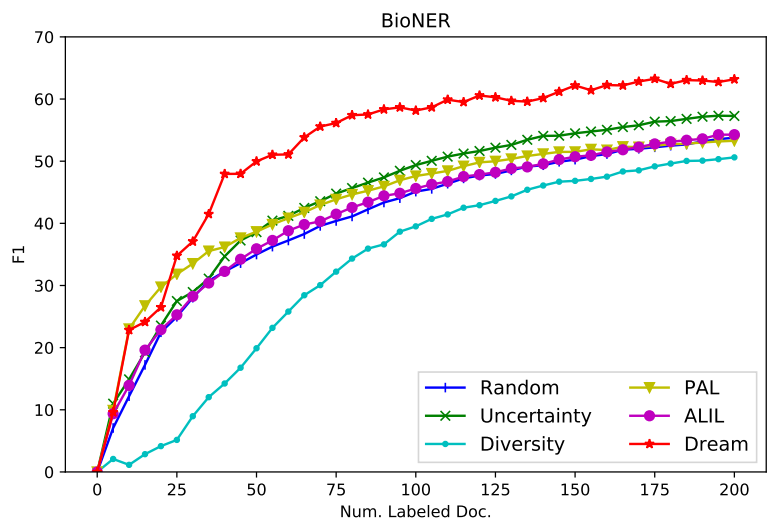}
    \caption{Top: Fig.~3 of \citep{fang2017learning}. Middle: Fig~.2 of \citep{liu2018learning}. Bottom: Fig.~4 of \citep{vu2019learning}. }
    \label{fig:meta-figs-3}
\end{figure}

For meta-active-learning methods, Fig.~3 of \citep{fang2017learning} (replicated as Fig.~\ref{fig:meta-figs-3} top) shows that the RL-based PAL (policy active learning) consistently outperforms uncertainty and random baselines in a cross-lingual transfer setting. However, Fig~.2 of \citep{liu2018learning} (replicated as Fig.~\ref{fig:meta-figs-3} middle) shows that PAL are hardly better than uncertainty or random baselines in both cross-domain and cross-lingual transfer settings, while their proposed ALIL (active learning by imitation learning) is better than all studied baselines. Nevertheless, Fig.~4 of \citep{vu2019learning} (replicated as Fig.~\ref{fig:meta-figs-3} bottom) again shows that both PAL and ALIL are not better than the uncertainty-based heuristic on a named entity recognition task, while their proposed active learning by dreaming method ranks the best. 

These results demonstrate concerning inconsistency and lack of generalization to even very similar datasets or tasks. In addition, the above-presented performance curves are typically the sole analysis conducted by the authors. Thus, we believe that it is enlightening to analyze more aspects of a proposed strategy and its performance. In addition, a comparison with the optimal strategy could reveal actionable ways to improve the proposed strategy, which is the main motivation of our work. 

\section{Considerations for Stochastic Acquisition Functions}
\label{app:stochastic-acq}
It is straightforward to extend the discussion in Sec.~\ref{problem-formulation} to stochastic acquisition functions, such as those that require stochastic uncertainty estimation, like BALD, or those that are represented as stochastic policies. 

First, we introduce $\zeta$ to capture such randomness, so that $\Delta\DL_{k+1}=\acq(\theta_{k}, \DL_k, \zeta)$ remains deterministic. Then we can establish a one-to-one relationship between $\acq(\cdot, \cdot, \zeta)$ and order $\sigma_\zeta$ on $\DU$. By definition of $\xi$-optimal order $\sigma_\xi^*$, we still have $q_{\xi}(\sigma_\xi^*)\geq q_{\xi}(\sigma_\zeta)$, and $q_{\xi}(\sigma_\xi^*)\geq \Exp[\zeta]{q_{\xi}(\sigma_\zeta)}$. In other words, the $\xi$-optimal order $\sigma_\xi^*$ outperforms both the expected quality of the stochastic acquisition function and any single instantiation of stochasticity. 

The above discussion suggests that not accounting or for stochasticity in acquisition function or unable to do so (i.e. using the expected quality) is inherently sub-optimal. Since many heuristics require uncertainty estimation through explicitly stochastic processes such as MC-Dropout \citep{gal2016dropout}, the stochastic components may become the bottleneck to further improvement as AL algorithms approach the optimal limit. 

\section{Model Architecture}
\label{app:arch}
\begin{table}[!htb]
    \centering
    \begin{subtable}[t]{0.24\textwidth}
    \resizebox{\textwidth}{!}{
    \begin{tabular}[t]{c}\toprule
        Input: $28 \times 28 \times 1$ \\\midrule
        Conv: $32\,\,\,5\times 5$ filters \\\midrule
        (Optional: MC-Dropout)\\\midrule
        Max-Pool: $2\times 2$\\\midrule
        ReLU\\\midrule
        Conv: $64\,\,\,5\times 5$ filters \\\midrule
        (Optional: MC-Dropout)\\\midrule
        Max-Pool: $2\times 2$\\\midrule
        ReLU\\\midrule
        Linear: $1024\times 128$\\\midrule
        (Optional: MC-Dropout)\\\midrule
        ReLU\\\midrule
        Linear: $128\times 10$\\\bottomrule
    \end{tabular}
    }
    \caption{CNN Object Class.}
    \end{subtable}
    \begin{subtable}[t]{0.24\textwidth}
    \resizebox{\textwidth}{!}{
    \begin{tabular}[t]{c}\toprule
        Input: $L$ tokens \\\midrule
        GloVe embedding: 300 dim\\\midrule
        BiLSTM: 300 hidden dim \\\midrule
        Average of $L$ hiddens\\\midrule
        (Optional: MC-Dropout)\\\midrule
        Linear: $600\times 7$\\\bottomrule
    \end{tabular}
    }
    \caption{LSTM Intent Class.}
    \end{subtable}
    \begin{subtable}[t]{0.24\textwidth}
    \resizebox{\textwidth}{!}{
    \begin{tabular}[t]{c}\toprule
        Input: $L$ tokens \\\midrule
        GloVe embedding: 300 dim\\\midrule
        Conv: 50 size-3 kernels \\\midrule
        ReLU\\\midrule
        Conv: 50 size-3 kernels \\\midrule
        ReLU\\\midrule
        Conv: 50 size-3 kernels \\\midrule
        ReLU\\\midrule
        Average-Pool\\\midrule
        Linear: $50\times 50$\\\midrule
        ReLU\\\midrule
        (Optional: MC-Dropout)\\\midrule
        Linear: $50\times 7$\\\bottomrule
    \end{tabular}
    }
    \caption{CNN Intent Class.}
    \end{subtable}
    \begin{subtable}[t]{0.24\textwidth}
    \resizebox{\textwidth}{!}{
    \begin{tabular}[t]{c}\toprule
        Input: $L$ tokens \\\midrule
        GloVe embedding: 300 dim\\\midrule
        Averege of $L$ embeddings \\\midrule
        Linear: $300\times 100$\\\midrule
        ReLU\\\midrule
        Linear: $100\times 100$\\\midrule
        ReLU\\\midrule
        (Optional: MC-Dropout)\\\midrule
        Linear: $100\times 7$\\\bottomrule
    \end{tabular}
    }
    \caption{AOE Intent Class.}
    \end{subtable}
    \caption{Model Architectures for Object and Intent Classification}
    \label{oc-ic-arch}
\end{table}

\subsection{Object Classification}
We follow the architecture used by \citet{kirsch2019batchbald}. The MC-Dropout layers are present in BALD and BatchBALD for uncertainty estimation. The architecture is summarized in Tab.~\ref{oc-ic-arch}(a). 

\subsection{Intent Classification}
We use an LSTM architecture as our main architecture of study for the intent classification task. We initialize the word embeddings with GloVe \citep{pennington2014glove} and jointly train it along with other model parameters. The architecture is shown in Tab.~\ref{oc-ic-arch}(b). Again, MC-Dropout is only used for uncertainty estimation in BALD. 

To study model transfer performance, we also used (1D-)CNN and Average of Embedding (AOE) architectures, shown in Tab.~\ref{oc-ic-arch}(c) and (d), as well as the pre-trained RoBERTa architecture \citep{liu2019roberta} with a linear classification layer on top. 

\subsection{Named Entity Recognition}
For named entity recognition, we used the same architecture as \citet{shen2017deep}, except that we removed character-level embedding layer since an ablation study did not find it beneficial to the performance. 

Specifically, the model architecture is encoder-decoder. The encoder builds the representation for each word in the same way as the BiLSTM architecture for intent classification. At each decoding step, the decoder maintains the running hidden and cell state, receives the concatenation of the current word encoder representation and previous tag one-hot representation as the input, and predicts the current tag with a linear layer from output hidden state. At the first step, a special \texttt{[GO]} tag was used. 

\section{Extended Quality Summary}
\label{app:qualities-full}
Tab.~\ref{tab:qualities-full} presents the numerical quality scores for optimal, heuristic, and random orders. The object and intent classification tasks are evaluated and averaged across 5 different model seeds, and the NER task is evaluated on a single model seed. In all cases, we can observe a large gap between the optimal order and the rest, indicating abundant room of improvement for the heuristics. 

\begin{table}[!htb]
    \centering
    \begin{tabular}{l|ll||l|l}\toprule
         & Object Class. & Intent Class. & & NER \\\midrule
         Optimal & 0.761 & 0.887 & Optimal & 0.838\\
         Max-Entorpy & 0.682 & 0.854 & Min-Confidence & 0.811\\
         BALD & 0.676 & 0.858 & Norm. Min-Conf. & 0.805\\
         BatchBALD & 0.650 & N/A & Longest & 0.793\\
         Random & 0.698 & 0.816 & Random & 0.801\\\bottomrule
    \end{tabular}
    \caption{Complete results of optimal, heuristic, and random order qualities. }
    \label{tab:qualities-full}
\end{table}

\section{Optimal Performance Curves}
\label{app:curves}
\begin{figure}[!htb]
    \centering
    \includegraphics[width=0.45\columnwidth]{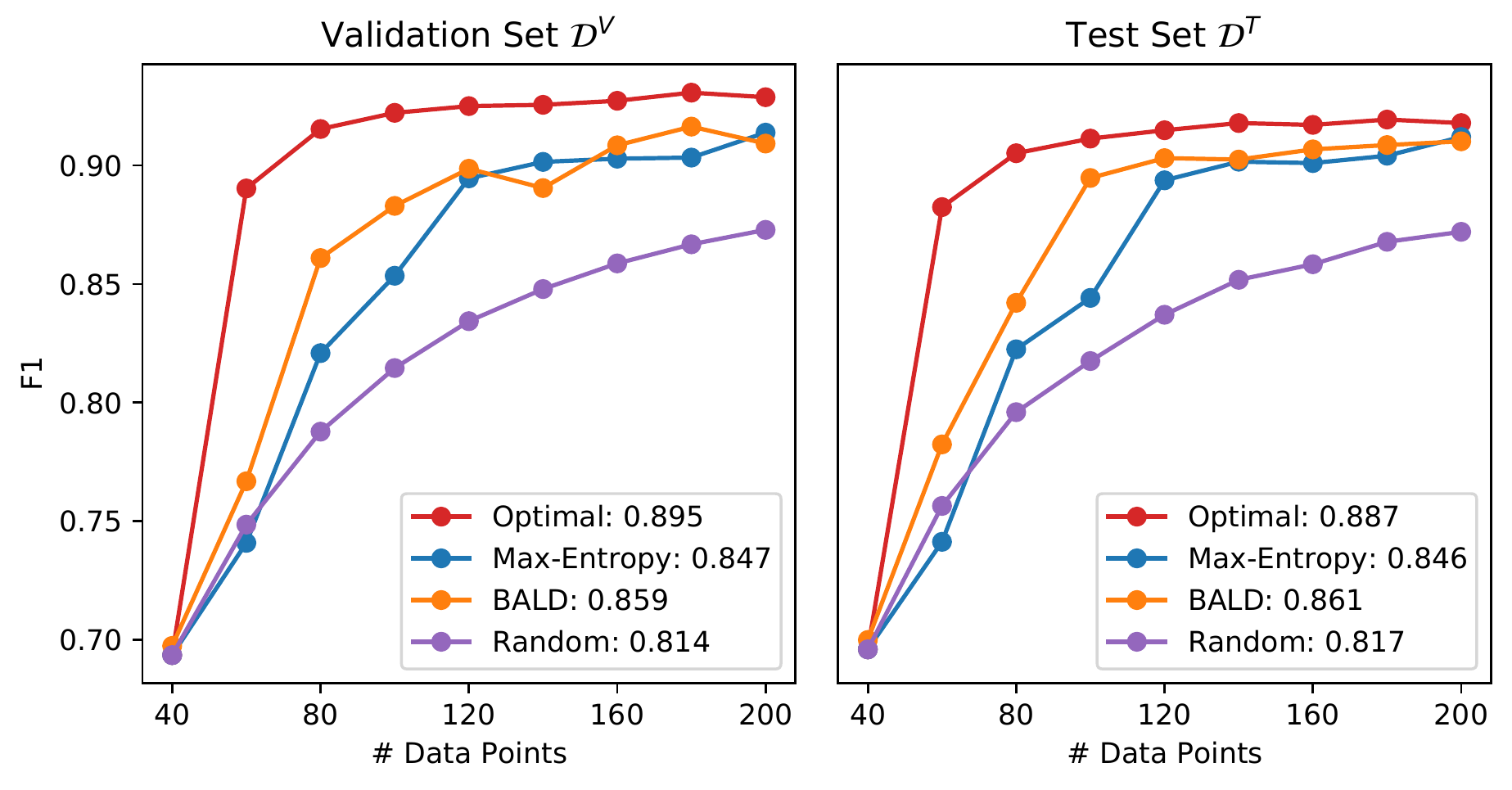}\hspace{0.3in}
    \includegraphics[width=0.45\columnwidth]{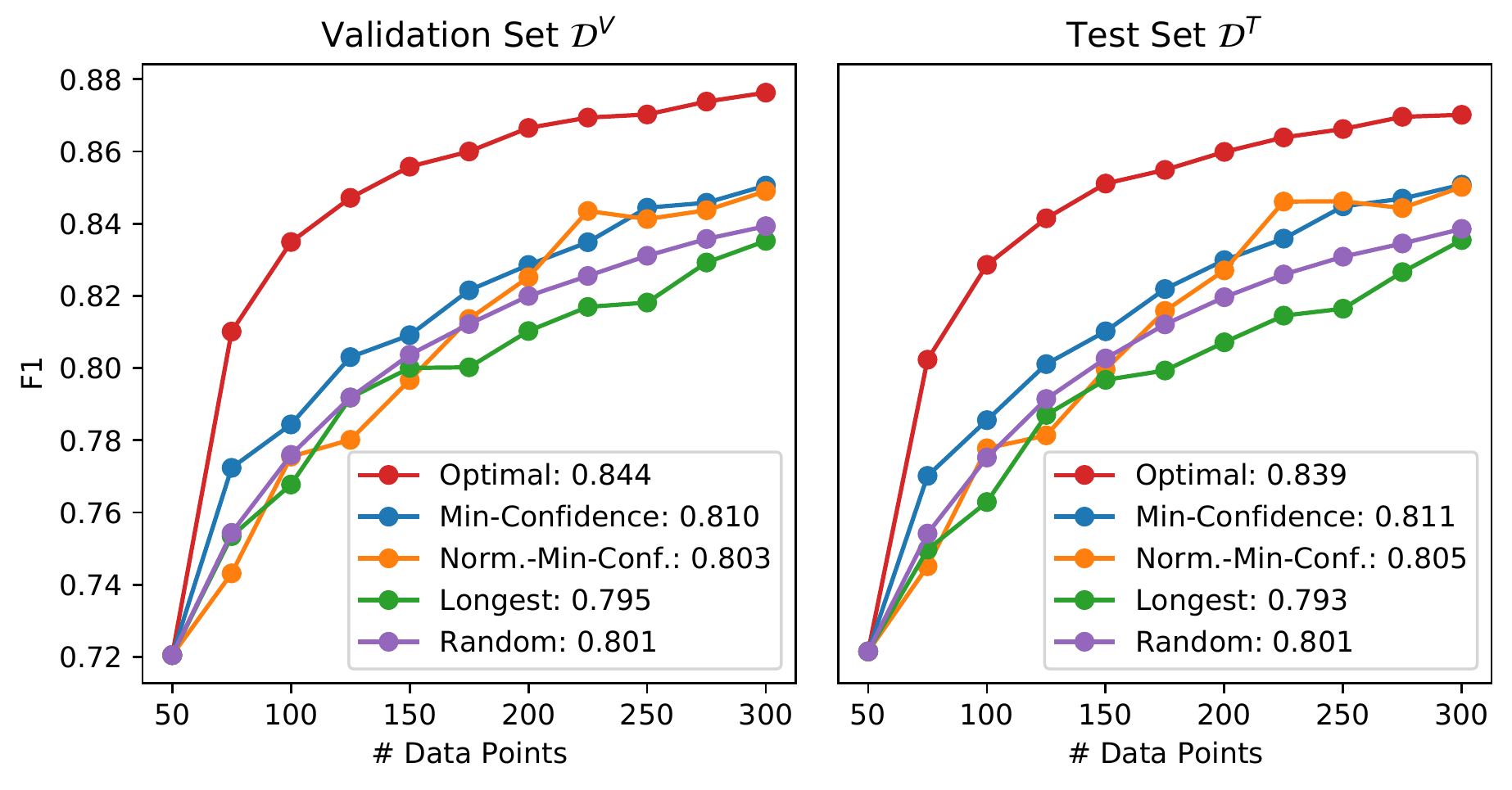}
    \caption{Performance curves for intent classification (left two panels) and NER (right two panels). }
    \label{fig:ic-ner-curves}
\end{figure}

Fig.~\ref{fig:ic-ner-curves} shows the performance curves on the validation and test sets for intent classification (left two panels) and named entity recognition (right two panels). Similar to those for object classification (Fig.~\ref{fig:oc-curves}), the shape of the curves are quite similar on validation and test sets, indicating good generalization. 

\section{Additional Visualizations for Object Classification}
\label{app:oc-distr}

\begin{figure}[!htb]
    \centering
    \includegraphics[width=0.85\columnwidth]{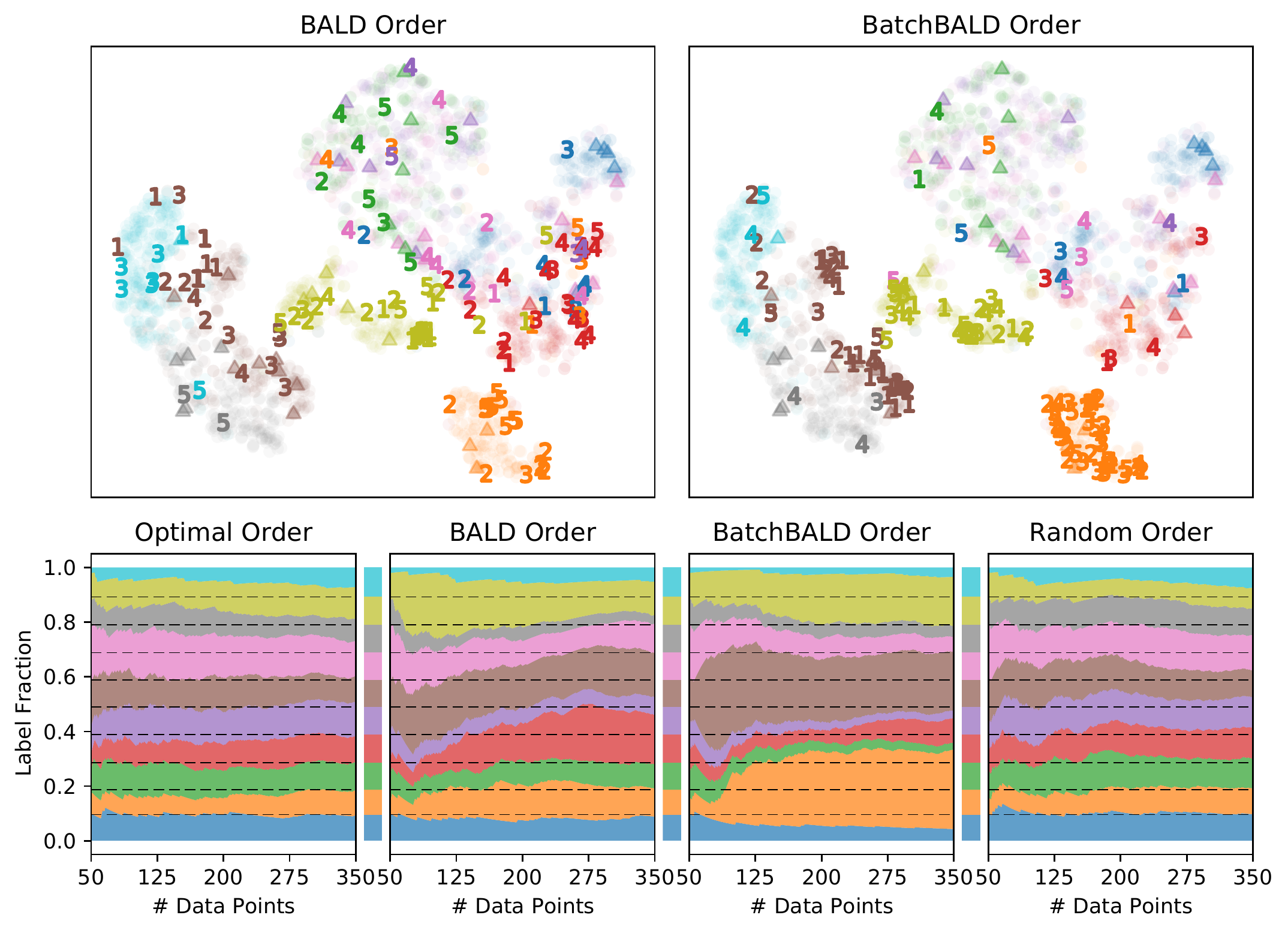}
    \caption{Top: the first five batches numerically labeled in t-SNE embedding space, along with warmstart (triangle) and test (circle) samples. Bottom: label distribution w.r.t. labeled set size, with test-set distribution shown between plots and as dashed lines. }
    \label{fig:bald-batchbald-distribution}
\end{figure}

Fig.~\ref{fig:bald-batchbald-distribution} presents the input (via PCA and t-SNE dimensionality reduction) and output distribution for BALD and BatchBALD order. The labels for data points sampled by the BALD and BatchBALD heuristics are much less balanced than the optimal and random orders. In addition, BatchBALD seems to focus on only a few concentrated regions in the input space. 

\section{Output Distribution-Matching Regularization (ODMR)}
\label{app:odmr}

In this section we describe the mixed results of regularizing the output distribution on object and intent classification. Since NER is a structured prediction task where the output tags in a sentence are correlated (e.g. I-tags always follow B- or I-tags), it is not clear what is the best way to enforce the distribution-matching property in the tag space and we leave the investigation to future work. 

Alg.~\ref{alg:odmr} presents the OMDR algorithm. Compared to the IDMR algorithm (Alg.~\ref{idmr}), there are two additional challenges. First, the labels for the pool set are not observed, so when the algorithm tries to acquire a data point with a certain label, it can only use the predicted label, which, with few data, can be very wrong in some cases. As soon as a data point is selected for annotation, its label is revealed immediately and used to calculate $d_{\mathrm{cur}}$ during the next data point selection. In other words, data annotations are not batched. 

Second, due to the unavailability of pool labels, the only labels for which we can use to compute the reference distribution are from $\DL_0$ and $\mathcal D^M$. If they are small, as are typically, the reference label distribution can be far from the true distribution $\mathbb P_Y$, especially for rare classes. Thus, we employ the add-1 smoothing (i.e. adding 1 to each label count before calculating the empirical count distribution, also known as Laplace smoothing) to ensure that rare classes are not outright missed. 

\begin{algorithm}[!htb]
\SetAlgoLined
    \SetKwInput{KwInput}{Input}
    \KwInput{$\acq\left(\DU, m_\theta, \mathcal D^L\right)$ that returns the next data point in $\DU$ to label}
    \hspace{-0.01in}\begin{tabular}{p{0.24cm}l}
    $ d_{\mathrm{ref}}$ & $=\texttt{label-distribution}\left(\DL_{0,Y}\cup \mathcal D^M_Y, \texttt{add-one=True}\right)$\;\\
    $ d_{\mathrm{cur}}$ & $= \texttt{label-distribution}\left(\mathcal D^L_Y, \texttt{add-one=False}\right)$\;\\
    $ l^*$ & $= \argmin_{l} \left(d_{\mathrm{cur}}-d_{\mathrm{ref}}\right)_l$\;\\
    $ \DU_{l^*}$ & $= \{x\in \DU: m_\theta(x)=l^*\}$\;\\
    \end{tabular}\\
    \Return $\acq\left(\DU_{b^*}, m_\theta, \mathcal D^L\right)$
\caption{Output Distribution-Matching Regularization (IDMR)}
\label{alg:odmr}
\end{algorithm}

In addition to Alg.~\ref{alg:odmr}, we also experiment with three ``cheating'' versions of ODMR. The most extreme version (test+groundtruth) uses test set (rather than the accessible set $\DL_0\cup \mathcal D^M$) to estimate label distribution and the groundtruth label (rather than the predicted label) are used to construct the subset $\mathcal D_{l^*}^U$. The other two versions (accessible+groundtruth and test+predicted) drops either piece of information. The actual ``non-cheating'' ODMR (accessible+predicted) is the only version that is practically feasible in reality. 

\begin{figure}[!htb]
    \centering
    \includegraphics[width=\columnwidth]{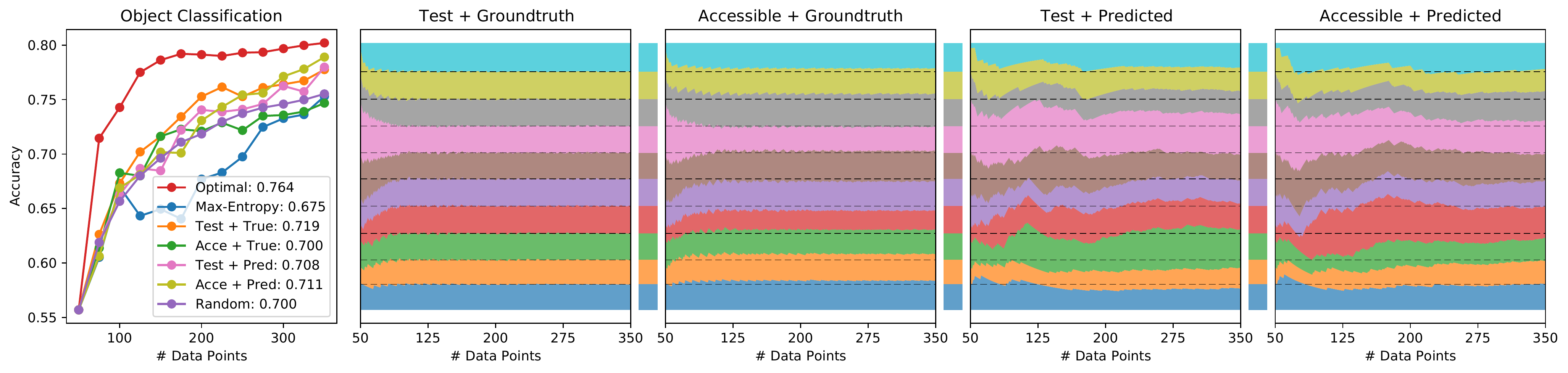}
    \caption{ODMR with max-entropy heuristic on object classification along with observed label distribution. }
    \label{fig:odmr-fmnist}
\end{figure}

Fig.~\ref{fig:odmr-fmnist} presents the four versions of ODMR-augmented heuristics on the intent classification result. We can see that all four versions are able to outperform the baseline heuristic, and three out of four (including the ``non-cheating'' ODMR) are able to outperform the random baseline. In addition, the label distribution are much more uniform compared to the unregularized version in Fig.~\ref{fig:fmnist-distribution} (bottom middle). 

\begin{figure}[!htb]
    \centering
    \includegraphics[width=\columnwidth]{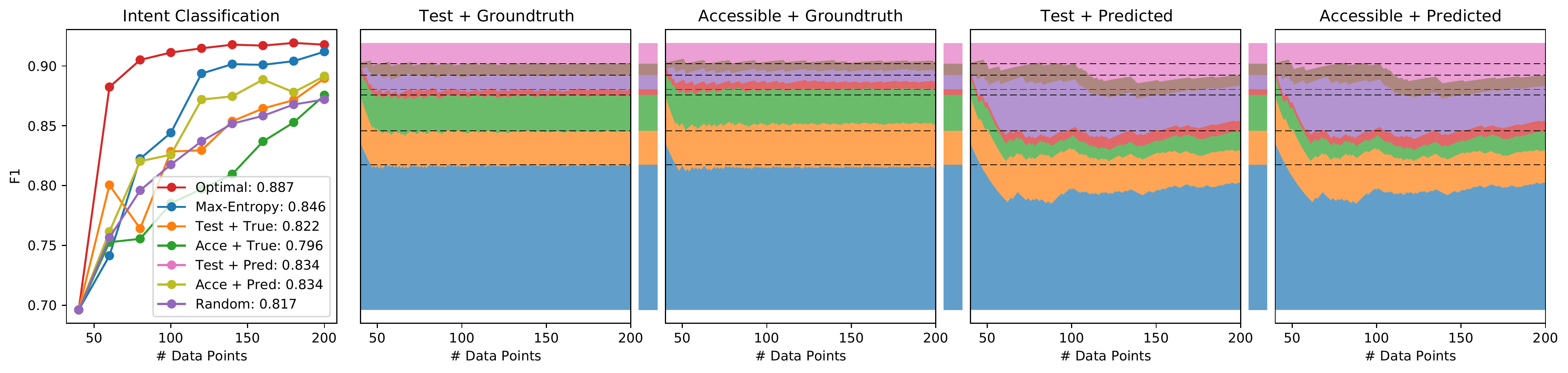}
    \includegraphics[width=\columnwidth]{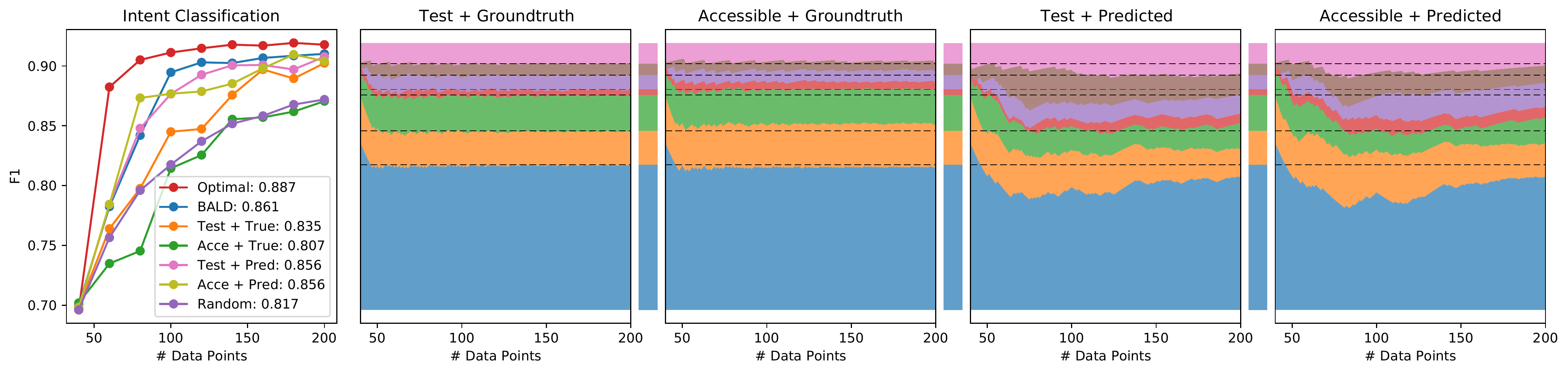}
    \caption{ODMR with max-entropy (top) and BALD (bottom) heuristics on intent classification along with observed label distribution. }
    \label{fig:odmr-ic}
\end{figure}

Fig.~\ref{fig:odmr-ic} presents the four versions of ODMR-augmented max-entropy (top) and BALD (bottom) heuristics on the intent classification result. Unlike object classification, none of the ``cheating'' or ``non-cheating'' versions are able to outperform the respective vanilla heuristic. Interestingly, using predicted rather than groundtruth labels achieves better performance in both cases.  

Overall, the results are inconclusive about whether ODMR can help. Since it uses the predicted label to guide data acquisition, this kind of bootstrapping is prone to ``confident mistakes'' when a data point is predicted wrong but with high certainty and thus never acquired, while the selection mechanism of IDMR provides a much more independent, yet still heuristic-guided, way to cover the data distribution. Visually, ODMR performance curves for both tasks are much less smooth than IDMR curves (Fig.~\ref{idmr-curves}), indicating unstability of using output labels to regularize a prediction task. Additional studies are required to determine when and how ODMR can help, and we leave them to future work.

\end{document}